\documentclass[aps,twocolumn,prx,superscriptadddegiuli2019randomress]{revtex4-2}
\usepackage[english]{babel}
\usepackage[utf8]{inputenc}
\usepackage{graphicx}
\usepackage{amsmath}
\usepackage{amsthm}
\usepackage{amssymb}
\usepackage{amsfonts}
\usepackage{mathtools}
\usepackage{enumerate}
\usepackage{bm}
\usepackage[dvipsnames]{xcolor}

\usepackage{appendix}
\usepackage{url}

\usepackage{hyperref}
\addto\extrasenglish{

    \def\equationautorefname~#1\null{Eq.~(#1)\null}

}
\newcommand{\suppfigref}[1]{\hyperref[#1]{Supplementary Fig.~\ref*{#1}}}

\hypersetup{
    colorlinks=true,
    linkcolor=blue,
    filecolor=magenta,      
    urlcolor=NavyBlue,
    citecolor=Blue
}

\newcommand{\avg}[1]{\left\langle #1 \right\rangle}

\newcommand{\cov}[1]{\text{Cov}\left[ #1 \right]}
\newcommand{\var}[1]{\text{Var}\left[ #1 \right]}
\newcommand{\prob}[1]{\text{Pr}\left\lbrace #1 \right\rbrace}
\newcommand{\Hg}[1]{\text{Hg}_{ #1}}

\newcommand{\be}{\begin{equation}}
\newcommand{\ee}{\end{equation}}
\newcommand{\ba}{\begin{eqnarray}}
\newcommand{\ea}{\end{eqnarray}}

\begin{document} 

\title{How Deep Neural Networks Learn Compositional Data:\\The Random Hierarchy Model}

\author{Francesco Cagnetta}
\thanks{These two authors contributed equally}
\affiliation{Institute of Physics, EPFL, Lausanne, Switzerland}
\author{Leonardo Petrini}
\thanks{These two authors contributed equally}
\affiliation{Institute of Physics, EPFL, Lausanne, Switzerland}
\author{Umberto M. Tomasini}
\affiliation{Institute of Physics, EPFL, Lausanne, Switzerland}
\author{Alessandro Favero}
\affiliation{Institute of Physics, EPFL, Lausanne, Switzerland}
\affiliation{Institute of Electrical Engineering, EPFL, Lausanne, Switzerland}
\author{Matthieu Wyart}
\email[Correspondence to ]{francesco.cagnetta@epfl.ch, matthieu.wyart@epfl.ch}
\affiliation{Institute of Physics, EPFL, Lausanne, Switzerland}

\begin{abstract}
Deep learning algorithms demonstrate a surprising ability to learn high-dimensional tasks from limited examples. This is commonly attributed to the depth of neural networks, enabling them to build a hierarchy of abstract, low-dimensional data representations. However, how many training examples are required to learn such representations remains unknown. To quantitatively study this question, we introduce the Random Hierarchy Model: a family of synthetic tasks inspired by the hierarchical structure of language and images. The model is a classification task where each class corresponds to a group of high-level features, chosen among several equivalent groups associated with the same class. In turn, each feature corresponds to a group of sub-features chosen among several equivalent ones and so on, following a hierarchy of composition rules. We find that deep networks learn the task by developing internal representations invariant to exchanging equivalent groups. Moreover, the number of data required corresponds to the point where correlations between low-level features and classes become detectable. Overall, our results indicate how deep networks overcome the \emph{curse of dimensionality} by building invariant representations, and provide an estimate of the number of data required to learn a hierarchical task.
\end{abstract}

\maketitle

Deep learning methods exhibit superhuman performances in areas ranging from image recognition~\cite{voulodimos2018deep} to Go-playing~\cite{silver2017mastering}. However, despite these accomplishments, we still lack a fundamental understanding of their working principles. Indeed, Go configurations and images lie in high-dimensional spaces, which are hard to sample due to the \emph{curse of dimensionality}: the distance $\delta$ between neighboring data points decreases very slowly with their number $P$, as $\delta =\mathcal{O}(P^{-1/d})$ where $d$ is the space dimension. Solving a generic task such as regression of a continuous function~\cite{luxburg2004distance} requires a small $\delta$, implying that $P$ must be {\it exponential} in the dimension $d$. Such a number of data is unrealistically large: for example, the benchmark dataset ImageNet~\cite{deng2009imagenet}, whose effective dimension is estimated to be $\approx 50$~\cite{pope2021intrinsic}, consists of only $\approx 10^7$ data, significantly smaller than $e^{50}\approx 10^{20}$. This immense difference implies that learnable tasks are not generic, but highly structured. What is then the nature of this structure, and why are deep learning methods able to exploit it?

A popular idea attributes the efficacy of these methods to their ability to build a useful representation of the data, which becomes increasingly complex across the layers~\cite{Lecun15}. Interestingly, a similar increase in complexity is also found in the visual cortex of the primate brain~\cite{van1983hierarchical,grill2004human}. In simple terms, neurons closer to the input learn to detect simple features like edges in a picture, whereas those deeper in the network learn to recognize more abstract features, such as faces~\cite{zeiler_visualizing_2014,doimo2020hierarchical}. Intuitively, if these representations are also invariant to aspects of the data unrelated to the task, such as the exact position of an object in a frame for image classification \cite{bruna2013invariant}, they may effectively reduce the dimensionality of the problem and make it tractable. This view is supported by several empirical studies of the hidden representations of trained networks. In particular, measures such as the mutual information between such representations and the input~\cite{shwartz2017opening,saxe2019information}, their intrinsic dimensionality~\cite{ansuini2019intrinsic,recanatesi2019dimensionality}, and their sensitivity toward transformations that do not affect the task (e.g., smooth deformations for image classification~\cite{petrini_relative_2021, tomasini_how_2022}), all eventually decay with the layer depth. However, none of these studies addresses the \emph{sample complexity}, i.e., the number of training data necessary for learning such representations, and thus the task.

In this paper, we study the relationship between sample complexity, depth of the learning method, and structure of the data by focusing on tasks with a hierarchically compositional structure---arguably a key property for the learnability of real data~\cite{patel2015probabilistic,mossel2016deep,mhaskar2016learning, poggio2017why, malach2018provably, demba19model, schmidt2020nonparametric,cagnetta2023can}. To provide a concrete example, consider a picture that consists of several high-level features like face, body, and background. Each feature is composed of sub-features like ears, mouth, eyes, and nose for the face, which can be further thought of as combinations of low-level features such as edges~\cite{grenander1996elements}. Recent studies have revealed that deep networks can represent hierarchically compositional functions with far fewer parameters than shallow networks~\cite{poggio2017why}, implying an information-theoretic lower bound on the sample complexity which is only \emph{polynomial} in the input dimension~\cite{schmidt2020nonparametric}. While these works offer important insights, they do not characterize the performance of deep neural networks trained with gradient descent. 

We investigate this question by adopting the physicist's approach~\cite{mezard2017mean, degiuli2019random, saxe2019mathematical,bahri2020statistical,ingrosso2022data} of introducing a model of synthetic data, which is inspired by the structure of natural problems, yet simple enough to be investigated systematically. This model (\autoref{sec:model}) belongs to a family of hierarchical classification problems where the class labels generate the input data via a hierarchy of composition rules. These problems were introduced to highlight the importance of input-to-label correlations for learnability~\cite{mossel2016deep} and were found to be learnable via an iterative clustering algorithm~\cite{malach2018provably}. Under the assumption of randomness of the composition rules, we show empirically that shallow networks suffer from the curse of dimensionality (\autoref{sec:empirical}), whereas the sample complexity $P^*$ of deep networks (both convolutional networks and multi-layer perceptrons) is only polynomial in the size of the input. More specifically, with $n_c$ classes and $L$ composition rules that associate $m$ equivalent low-level representations to each class/high-level features, $P^* \simeq n_c m^L$ asymptotically in $m$ (\autoref{sec:empirical}). 

Furthermore, we find that $P^*$ coincides with both \emph{(a)} the number of data that allows for learning a representation that is invariant to exchanging the $m$ semantically equivalent low-level features~(\autoref{ssec:syninv}) and \emph{(b)} the size of the training set for which the correlations between low-level features and class label become detectable~(\autoref{sec:correlations}). We prove for a simplified architecture trained with gradient descent that \emph{(a)} and \emph{(b)} must indeed coincide. Via \emph{(b)}, $P^*$ can be derived analytically under our assumption of randomness of the composition rules.

\subsection{Relationship to other models of data structure}

Characterizing the properties that make high-dimensional data learnable is a classical problem in statistics. Typical assumptions that allow for avoiding the curse of dimensionality include \textit{(i)} data lying on a low-dimensional manifold and \textit{(ii)} the task being smooth~\cite{bach2021quest}. For instance, in the context of regression, the sample complexity is not controlled by the bare input dimensionality $d$, but by the ratio $d_M/s$~\cite{gyorfi2002distribution, kpotufe2011k, hamm2021adaptive}, where $d_M$ is the dimension of the data manifold and $s$ the number of bounded derivatives of the target function. However, $d_M$ is also large in practice~\cite{pope2021intrinsic}, thus keeping $d_M/s$ low requires an unrealistically large number of bounded derivatives. Moreover, properties \textit{(i)} and \textit{(ii)} can already be leveraged by isotropic kernel methods, and thus cannot account for the significant advantage of deep learning methods in many benchmark datasets~\cite{geiger2020perspective}. Alternatively, learnability can be achieved when \textit{(iii)} the task depends on a small number of linear projections of the input variables, such as regression of a target function $f^*(x)=g(x_t)$ where $x \in \mathbb{R}^d$ and $x_t \in \mathbb{R}^t$ \cite{paccolat2020compressing,abbe2021staircase,barak2022hidden,dandi2023two}. Methods capable of learning features from the data can leverage this property to achieve a sample complexity that depends on $t$ instead of $d$~\cite{bach2017breaking}. However, one-hidden-layer networks are sufficient for that, hence this property does not explain the need for deep architectures.

In the context of statistical physics, the quest for a model of data structure has been pursued within the framework of teacher-student models~\cite{gardner1989three, zdeborova2016statistical, mezard2023spin}, where a teacher uses some ground truth knowledge to generate data, while a student tries to infer the ground truth from the data. The structural properties \textit{(i,ii,iii)} can be incorporated into this approach~\cite{spigler2019asymptotic, goldt2019modelling}. In addition, using a shallow convolutional network as a teacher allows for modeling \textit{(iv)} the \emph{locality} of image-like datasets~\cite{favero2021locality, ingrosso2022data, aiudi2023local}. In the context of regression, this property can be modelled with a function $f^*(x)=\sum_i f_i^*(x_i)$ where the sum is on all patches $x_i$ of $t$ adjacent pixels. Convolutional networks learn local tasks with a sample complexity controlled by the patch dimension $t$~\cite{favero2021locality}, even in the `lazy' regime \cite{jacot2018neural,chizat2019lazy} where they do not learn features. However, locality does not allow for long-range nonlinear dependencies in the task. It might be tempting to include these dependencies by considering a deep convolutional teacher network, but then the sample complexity would be exponential in the input dimension $d$~\cite{cagnetta2023can}.

The present analysis based on hierarchical generative models shows that properties \textit{(i,ii,iii)} are not necessary to beat the curse of dimensionality. Indeed, for some choices of the parameters, the model generates all possible $d$-dimensional sequences of input features, which violates \textit{(i)}. Additionally, changing a single input feature has a finite probability of changing the label, violating the smoothness assumption \textit{(ii)}. Finally, the label depends on all of the $d$ input variables of the input, violating \textit{(iii)}. Yet, we find that the sample complexity of deep neural networks is only polynomial in $d$. Since locality is incorporated hierarchically in the generative process, it generates long-range dependencies in the task, but it can still be leveraged by building a hierarchical representation of the data.
\begin{figure*}
    \centering
    \includegraphics[width=\linewidth]{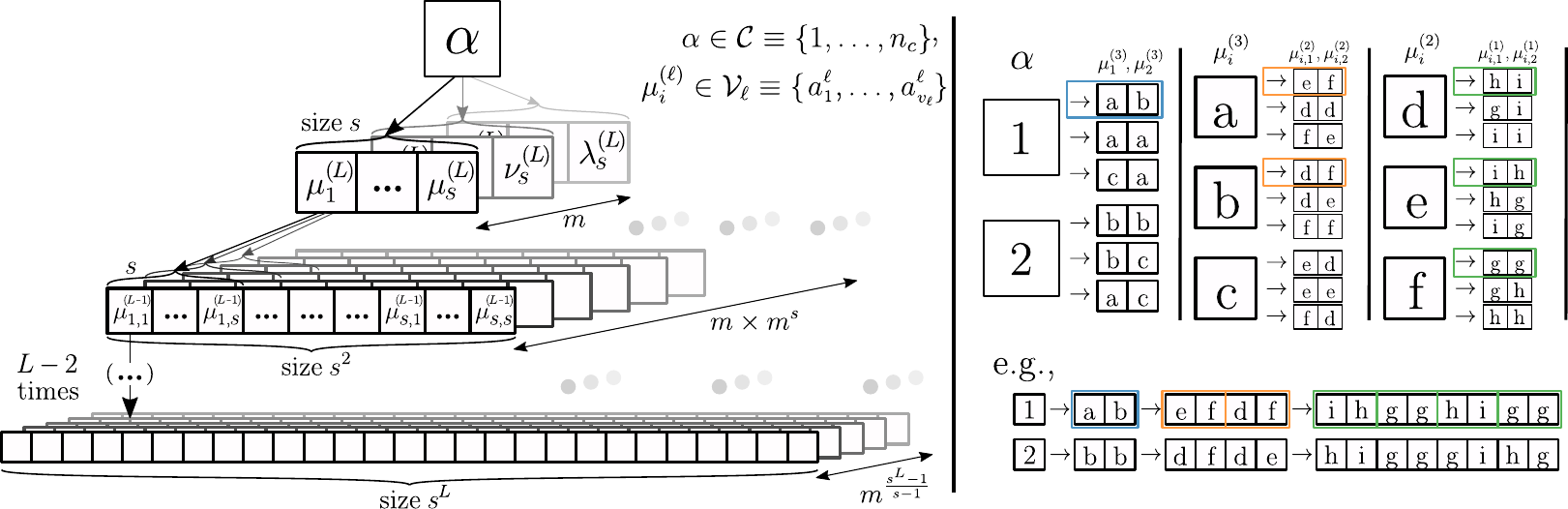}
    \caption{\textbf{The Random Hierarchy Model. Left:} Structure of the generative model. The class label $\alpha\,{=}\,1,\dots,n_c$ generates a set of $m$ equivalent (i.e., \emph{synonymic}) high-level representations with elements taken from a vocabulary of high-level features $\mathcal{V}_L$. Similarly, high-level features generate $m$ equivalent lower-level representations, taken from a vocabulary $\mathcal{V}_{L-1}$. Repeating this procedure $L\,{-}\,2$ times yields all the input data with label $\alpha$, consisting of low-level features taken from $\mathcal{V}_1$. \textbf{Right:} example of Random Hierarchy Model with $n_c\,{=}\,2$ classes, $L\,{=}\,3$, $s\,{=}\,2$, $m\,{=}\,3$ and homogeneous vocabulary size $v_1\,{=}\,v_2\,{=}\,v_3\,{=}\,3$. The three sets of rules are listed at the top, while two examples of data generation are shown at the bottom. The first example is obtained by following the rules in the colored boxes.}
    \label{fig:data_net_illustration}
\end{figure*} 

\section{The Random Hierarchy Model}\label{sec:model}

In this section, we introduce our generative model, which can be thought of as an $L$-level context-free grammar---a generative model of language from formal language theory~\cite{rozenberg_handbook_1997}. The model consists of a set of class labels $\mathcal{C}\equiv\left\lbrace 1,\dots, n_c\right\rbrace$ and $L$ disjoint vocabularies $\mathcal{V}_\ell\,{\equiv}\,\left\lbrace a^{\ell}_1,\dots,a^{\ell}_{v_\ell}\right\rbrace$ of low- and high-level features. As illustrated in~\autoref{fig:data_net_illustration}, left panel, data are generated from the class labels. Specifically, each label generates $m$ distinct high-level representations via $m$ composition rules of the form
\begin{equation}
\alpha \mapsto \mu^{(L)}_1,\dots,\mu^{(L)}_s\qquad\text{ for }\alpha\in\mathcal{C}\text{ and }\mu^{(L)}_i \in \mathcal{V}_L,
\end{equation}
having size $s\,{>}\,1$. The $s$ elements of these representations are high-level features $\mu^{(L)}_i$ such as background, face, and body for a picture. Each high-level feature generates in turn $m$ lower-level representations via other $m$ rules,
\begin{equation}
\mu^{(\ell)} \mapsto \mu^{(\ell-1)}_{1},\dots,\mu^{(\ell-1)}_{s}\text{ for }\mu^{(\ell)}\in\mathcal{V}_{\ell},\mu^{(\ell-1)}_{i} \in \mathcal{V}_{\ell-1},
\end{equation}
from $\ell\,{=}\,L$ down to $\ell\,{=}\,1$. The input features $\mu^{(1)}$ represent low-level features such as the edges in an image. Due to the hierarchical structure of the generative process, each datum can be represented as a tree of branching factor $s$ and depth $L$, where the root is the class label, the leaves are the input features, and the hidden nodes are the level-$\ell$ features with $\ell\,{=}\,2,\dots, L$.

In addition, for each level $\ell$, there are $m$ distinct rules emanating from the same higher-level feature $\mu^{(\ell)}$, i.e., there are $m$ equivalent lower-level representations of $\mu^{(\ell)}$ (see~\autoref{fig:data_net_illustration}, right panel, for an example with $m\,{=}\,3$). Following the analogy with language, we refer to these equivalent representations as \emph{synonyms}. We assume that a single low-level representation can only be generated by one high-level feature, i.e., that there are no ambiguities. Since the number of distinct $s$-tuples at level $\ell$ is bounded by $v_{\ell}^s$, this assumption requires $m v_{\ell+1} \,{\leq}\, v_{\ell}^{s}$ for all $\ell\,{=}\,1,\dots, L$ (with $v_{L+1}\,{\equiv}\,n_c$). If $m\,{=}\,1$, each label generates only a single datum and the model is trivial. For $m\,{>}\,1$, the number of data per class grows exponentially with the input dimension $d\,{=}\,s^L$,
\begin{equation}\label{eq:data-per-class}
m\times m^{s} \times \dots \times m^{s^{L-1}} = m^{ \sum_{i=0}^{L-1} s^i } = m^{\frac{d-1}{s-1}}.
\end{equation}
In particular, in the case where $m v_{\ell+1} \,{=}\, v_{\ell}^{s}$, the model generates all the possible data made of $d$ features in $\mathcal{V}_1$. Instead, for $m v_{\ell+1} \,{<}\, v_{\ell}^{s}$, the set of available input data is given by the application of the composition rules, therefore it inherits the hierarchical structure of the model.

Let us remark that, due to the non-ambiguity assumption, each set of composition rules can be summarized with a function $g_\ell$ that associates $s$-tuples of level-$\ell$ features to the corresponding level-$(\ell+1)$ feature. The domain of $g_\ell$ is a subset of $\mathcal{V}_{\ell}^s$ consisting of the $m v_{\ell+1}$ $s$-tuples generated by the features at level $(\ell+1)$. Using these functions, the label $\alpha\equiv \mu^{(L+1)}$ of an input datum $\bm{\mu}^{(1)}\,{=}\,\left(\mu_1^{(1)}, \dots,\mu_{d}^{(1)} \right)$ can be written as a hierarchical composition of $L$ local functions of $s$ variables~\cite{mhaskar2016learning, poggio2017why}:
\begin{equation}\label{eq:compositionality}
\mu_i^{(\ell+1)} = g_\ell \left( \mu^{(\ell)}_{(i-1)s+1},\dots, \mu^{(\ell)}_{(i-1)s+1}\right),
\end{equation}
for $i\,{=}\,1,\dots,s^{L-\ell}$ and $\ell\,{=}\,1,\dots,L$.

Notice that, while we keep $s$ and $m$ constant throughout the levels for ease of exposition, our results can be generalized without additional effort. Likewise, we will set the vocabulary size to $v$ for all levels. To sum up, a single classification task is specified by the parameters $n_c$, $v$, $m$ and $s$ and by the $L$ composition rules. In the \emph{Random Hierarchy Model} (RHM) the composition rules are chosen uniformly at random over all the possible assignments of $m$ representations of $s$ low-level features to each of the $v$ high-level features. An example of binary classification task ($n_c\,{=}\,2$), with $s\,{=}\,2$, $L\,{=}\,3$, and $v\,{=}\,m\,{=}\,3$, is shown in~\autoref{fig:data_net_illustration}, right panel, together with two examples of label-input pairs. Notice that the random choice induces correlations between low- and high-level features. In simple terms, each of the high-level features---e.g., the level-$2$ features $d$, $e$ or $f$ in the figure---is more likely to be represented with a certain low-level feature in a given position---e.g., $i$ on the right for $d$, $g$ on the right for $e$ and $h$ on the right for $f$. These correlations are crucial for our predictions and are analyzed in detail in~\autoref{app:correlations}.

\section{Sample Complexity of Deep Neural Networks}\label{sec:empirical}

The main focus of our work is the answer to the following question.
\begin{itemize}
\item[\textbf{Q:}] {\it How much data is required to learn a typical instance of the Random Hierarchy Model with a deep neural network?}
\end{itemize}
Thus, after generating an instance of the RHM with fixed parameters $n_c$, $s$, $m$, $v$, and $L$, we train neural networks of varying depth with stochastic gradient descent (SGD) on a set of $P$ training points. The training points are sampled uniformly at random without replacement from the set of available RHM data, hence they are all distinct. We adopt a one-hot encoding of the input features, so that each input point $\bm{x}$ is a $d\times v$-dimensional sequence where, for $i\,{=}\,1,\dots,d$ and $\nu\in\mathcal{V}_1$,
\begin{equation}
x_{i,\nu} = \left\lbrace\begin{aligned}1,& \text{ if }\mu^{(1)}_i=\nu,\\ 0, & \text{ otherwise.}\end{aligned}\right.
\end{equation}
All our experiments consider over-parameterized networks, which we achieve in practice by choosing the width $H$ of the network's hidden layers such that \emph{i)} training loss reaches $0$ \emph{ii)} test accuracy does not improve by increasing $H$. To guarantee representation learning as $H$ grows, we consider the maximal update parametrization~\cite{yang2020feature}, equivalent to having the standard ${H}^{-1/2}$ scaling of the hidden layer weights plus an extra factor of ${H}^{-1/2}$ at the last layer. Further details of the machine learning methods can be found in~\autoref{sec:matmet}.

\begin{figure}[h]
    \centering
    \hspace*{-.6cm}
    \includegraphics[width=.9\linewidth]{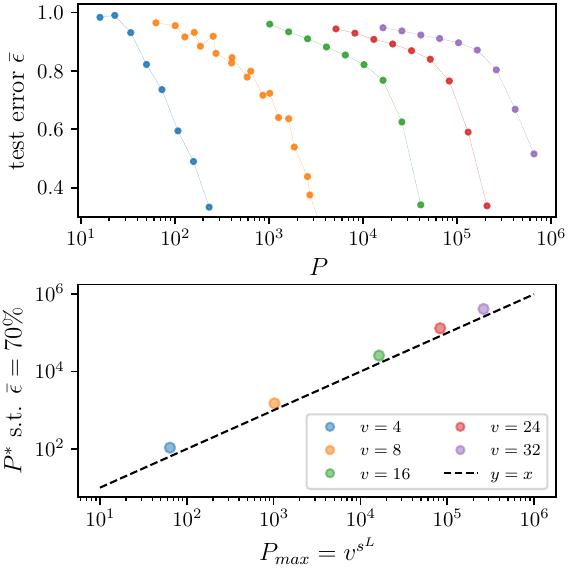}
    \caption{\textbf{Sample complexity of two-layer fully-connected networks, for $L=s\,{=}\,2$ and $v\,{=}\,n_c\,{=}\,m$.} Top: Test error vs the number of training data. Different colors correspond to different vocabulary sizes $v$. Bottom: number of training data resulting in test error $\bar{\epsilon}\,{=}\,0.7$ as a function of $P_{\text{max}}$, with the black dashed line indicating a linear relationship.}    
    \label{fig:terr_vs_p_ohl}
\end{figure}

\paragraph{Shallow networks are cursed.} Let us begin with the sample complexity of two-layer fully-connected networks. As shown in~\autoref{fig:terr_vs_p_ohl}, in the maximal case $n_c\,{=}\,v$, $m\,{=}\,v^{s-1}$ these networks
learn the task only if trained on a significant fraction of the total number of data $P_{\text{max}}$. From~\autoref{eq:data-per-class},
\begin{equation}\label{eq:num_data}
P_{\rm max} = n_c m^{\frac{d-1}{s-1}},
\end{equation}
which equals $v^{s^L}$ in the maximal case. The bottom panel of~\autoref{fig:terr_vs_p_ohl}, in particular, highlights that the number of training data required for having a test error $\epsilon\,{\leq}\,0.7\,\epsilon_{\text{rand}}$, with $\epsilon_{\text{rand}}\,{=}\,1-n_c^{-1}$ denoting the error of a random guess of the label, is proportional to $P_{\text{max}}$. Since $P_{\text{max}}$ is exponential in $d$, this is an instance of the curse of dimensionality.

\begin{figure}[h]
    \centering
    \hspace*{-.6cm}
    \includegraphics[width=.9\linewidth]{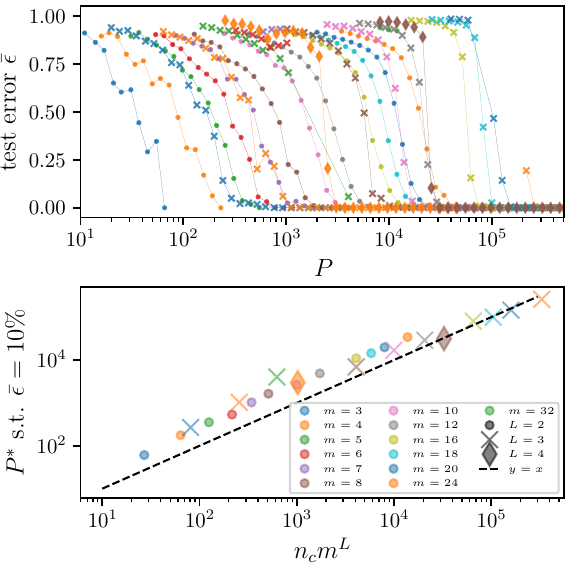}
    \caption{\textbf{Sample complexity of depth-$(L+1)$ CNNs,  for $s\,{=}\,2$ and $m\,{=}\,n_c\,{=}\,v$.} Top: Test error vs number of training points. Different colors correspond to different vocabulary sizes $v$ while the markers indicate the hierarchy depth $L$. Bottom: sample complexity $P^*$ corresponding to a test error $\epsilon^*=0.1 \epsilon_{\text{rand}}$. The empirical points show remarkable agreement with the law $P^* = n_c m^{L}$, shown as a black dashed line.}
    \label{fig:terr_vs_p_cnn}
\end{figure}

\begin{figure}[h]
    \centering
    \hspace*{-.6cm}
    \includegraphics[width=.9\linewidth]{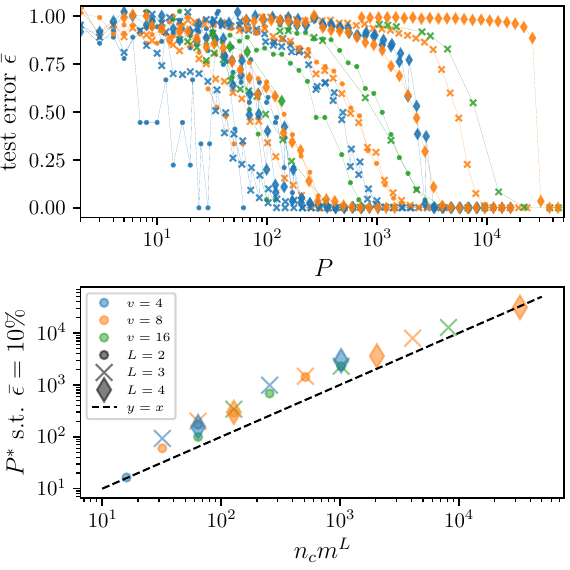}
    \caption{\textbf{Sample complexity of depth-$(L+1)$ CNNs, for $s\,{=}\,2$, $n_c\,{=}\,v$ and varying $m\,{\leq}\,v$.} Top: Test error vs number of training points, with different colors corresponding to different vocabulary sizes $v$ and markers indicating the hierarchy depth $L$. Bottom: sample complexity $P^*$, with the law $P^* = n_c m^L$ shown as a black dashed line.}
    \label{fig:testerror_vs_p_mln}
\end{figure}

\paragraph{Deep networks break the curse.} For networks having a depth larger than that of the RHM $L$, the test error displays a sigmoidal behavior as a function of the training set size. This finding is illustrated in the top panels of~\autoref{fig:terr_vs_p_cnn} and~\autoref{fig:testerror_vs_p_mln} (and~\autoref{fig:terr_vs_p_diff_nc} of~\autoref{app:add-figures} for varying $n_c$) for Convolutional Neural Networks (CNNs) of depth $L+1$ (details in~\autoref{sec:matmet}). Similar results are obtained for multi-layer perceptions of depth $\,{>}\,L$, as shown in~\autoref{app:add-figures}. All these results suggest the existence of a well-defined number of training data at which the task is learned. Mathematically, we define the sample complexity $P^*$ as the smallest training set size $P$ such that the test error $\epsilon(P)$ is smaller than $\epsilon_{\text{rand}}/10$. The bottom panels of~\autoref{fig:terr_vs_p_cnn} and~\autoref{fig:testerror_vs_p_mln} (and~\autoref{fig:terr_vs_p_diff_nc}, \autoref{fig:terr_vs_p_diff_arch}) show that
\begin{equation}\label{eq:empirical-sc}
P^* \simeq n_c m^L \Leftrightarrow \frac{P^*}{n_c}\simeq d^{\,\ln(m)/\ln(s)},
\end{equation}
independently of the vocabulary size $v$. Since $P^*$ is a power of the input dimension $d\,{=}\,s^L$, the curse of dimensionality is beaten, which evidences the ability of deep networks to harness the hierarchical compositionality of the task. It is crucial to note, however, that this ability manifests only in feature learning regimes, e.g., under the maximal update parameterization considered in this work. Conversely, as shown in~\autoref{fig:ntk} of~\autoref{app:add-figures} for the maximal case $n_c\,{=}\,v$, $m\,{=}\,v^{s-1}$, deep networks trained in the `lazy' regime \cite{jacot2018neural}---where they do not learn features---suffer from the curse of dimensionality, even when their architecture is matched to the structure of the RHM.

We now turn to study the internal representations of trained networks and the mechanism that they employ to solve the task.

\subsection{Emergence of Synonymic Invariance in Deep CNNs}\label{ssec:syninv} 

A natural approach to learning the RHM would be to identify the sets of $s$-tuples of input features that correspond to the same higher-level feature, i.e., synonyms. Identifying synonyms at the first level would allow for replacing each $s$-dimensional patch of the input with a single symbol, reducing the dimensionality of the problem from $s^L$ to $s^{L-1}$. Repeating this procedure $L$ times would lead to the class labels and, consequently, to the solution of the task. 

To test if deep networks trained on the RHM resort to a similar solution, we introduce the \emph{synonymic sensitivity}, which is a measure of the invariance of a function with respect to the exchange of synonymic low-level features. Mathematically, we define $S_{k,l}$ as the sensitivity of the $k$-th layer representation of a deep network with respect to exchanges of synonymous $s$-tuples of level-$l$ features. Namely,
\begin{equation}
\label{eq:stability-def}
    S_{k, l} = \frac{\langle\|f_{k}(\bm{x}) - f_{k}(P_l \bm{x})\|^2 \rangle_{\bm{x}, P_l}}{\langle\|f_{k}(\bm{x}) - f_{k}(\bm{y})\|^2\rangle_{\bm{x}, \bm{y}}},
\end{equation}
where: $f_k$ is the sequence of activations of the $k$-th layer in the network; $P_l$ is an operator that replaces all the level-$l$ tuples with one of their $m-1$ synonyms chosen uniformly at random; $\langle\cdot\rangle$ with subscripts $\bm{x},\bm{y}$ denotes average over pairs of input data of an instance of the RHM; the subscript $P_l$ denotes average over all the exchanges of synonyms.

\begin{figure}[h]
    \centering
    \includegraphics[width=0.9\linewidth]{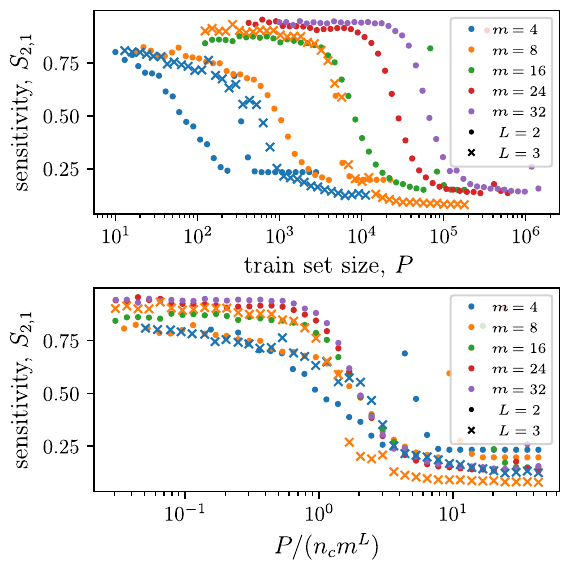}
    \caption{\textbf{Synonymic sensitivity $S_{2, 1}$ for a depth-$(L+1)$ CNN trained on the RHM with $s=2$, $n_c=m=v$} as a function of the training set size ($L$ and $v$ as in the key). The collapse achieved after rescaling by $P^* = n_c m^L$ highlights that the sample complexity coincides with the number of training points required to build internal representations invariant to exchanging synonyms.}
    \label{fig:sensitivity_collapse}
\end{figure}

\autoref{fig:sensitivity_collapse} reports $S_{2,1}$, which measures the sensitivity to exchanges of synonymic tuples of input features, as a function of the training set size $P$ for Deep CNNs trained on RHMs with different parameters. We focused on $S_{2,1}$---the sensitivity of the second layer of the network---since a single linear transformation of the input cannot produce an invariant representation in general.~\footnote{Let us focus on the first $s$-dimensional patch of the input $\bm{x}_1$, which can take $mv$ distinct values---$m$ for each of the $v$ level-$2$ features. For a linear transformation, insensitivity is equivalent to the following set of constraints: for each level-$2$ features $\mu$, and $\bm{x}_{1,i}$ encoding for one of the $m$ level-$1$ representations generated by $\mu$, $\bm{w}\cdot\bm{x}_{1,i}\,{=}\,c_{\mu}$. Since $c_\mu$ is an arbitrary constant, there are $v\times(m-1)$ constraints for the $v\times s$ components of $\bm{w}$, which cannot be satisfied in general unless $m\leq(s+1)$.} Notice that all the curves display a sigmoidal shape, signaling the existence of a characteristic sample size which marks the emergence of synonymic sensitivity in the learned representations. Remarkably, by rescaling the $x$-axis by the sample complexity of~\autoref{eq:empirical-sc} (bottom panel), curves corresponding to different parameters collapse. We conclude that the generalization ability of a network relies on the synonymic invariance of its hidden representations.

\begin{figure}[h]
    \centering
    \includegraphics[width=\linewidth]{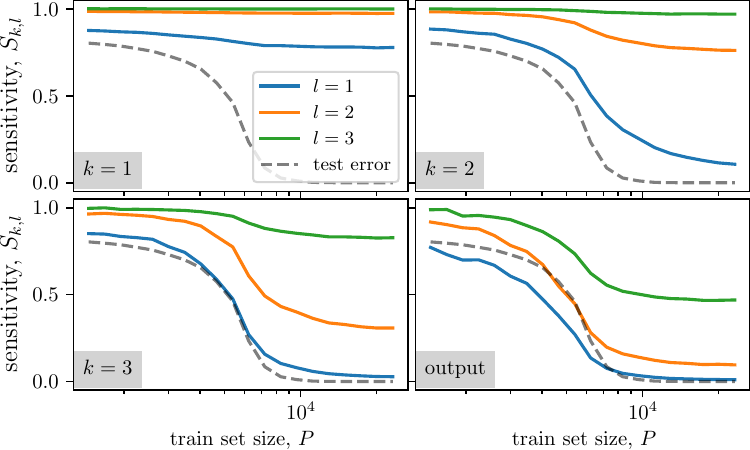}
    \caption{\textbf{Synonymic sensitivities $S_{k, l}$ of the layers of a depth-$(L+1)$ CNN trained on a RHM with $L\,{=}\,3$, $s\,{=}\,2$, $n_c\,{=}\,m\,{=}\,v\,{=}\,8$}, as a function of the training set size $P$. The colors denote the level of the exchanged synonyms (as in the key), whereas different panels correspond to the sensitivity of the activations of different layers (layer index in the gray box). Synonymic invariance is learned at the same training set size for all layers, and invariance to level-$l$ exchanges is obtained from layer $k=l+1$.}
    \label{fig:sensitivity_vs_P_bylayer}
\end{figure}

Measures of the synonymic sensitivity $S_{k,1}$ for different layers $k$ are reported in \autoref{fig:sensitivity_vs_P_bylayer} (blue lines), showing indeed that the layers $k\,{\geq}\,2$ become insensitive to exchanging level-$1$ synonyms. \autoref{fig:sensitivity_vs_P_bylayer} also shows the sensitivities to exchanges of higher-level synonyms: all levels are learned together as $P$ increases, and invariance to level-$l$ exchanges is achieved from layer $k=l+1$. The test error is also shown (gray dashed) to further emphasize its correlation with synonymic invariance.

\paragraph{Synonymic invariance and effective dimension.} Notice that the collapse of the representations of synonymic tuples to the same value implies a progressive reduction of the effective dimensionality of the hidden
representations, as reported in~\autoref{fig:deff} of~\autoref{app:intrinsic-dimension}.

\section{Correlations Govern Synonymic Invariance}\label{sec:correlations}

We now provide a theoretical argument for understanding the scaling of $P^*$ of~\autoref{eq:empirical-sc} with the parameters of the RHM.
First, we compute a third characteristic sample size $P_c$, defined as the size of the training set for which the {\it local} correlations between any of the input patches and the label become detectable. Remarkably, $P_c$ coincides with $P^*$ of~\autoref{eq:empirical-sc}. Secondly, we demonstrate how a shallow (two-layer) neural network acting on a single patch can use such correlations to build a synonymic invariant representation in a single step of gradient descent so that $P_c$ and $P^*$ also correspond to the emergence of an invariant representation. Lastly, we show empirically that removing such correlations leads again to the curse of dimensionality, even if the network architecture is matched to the structure of the RHM.

\subsection{Identify Synonyms by Counting}
\label{subsec:counting}
Groups of input patches forming synonyms can be inferred by counting, at any given location, the occurrences of such patches in all the data corresponding to a given class $\alpha$. Indeed, tuples of features that appear with identical frequencies are likely synonyms. More specifically, let us denote $\bm{x}_j$ an $s$-dimensional input patch  for $j$ in $1,\dots, s^{L-1}$, a $s$-tuple of input features with $\bm{\mu}\,{=}\,(\mu_1,\dots,\mu_s)$, and the number of data in class $\alpha$ having $\bm{x}_j=\bm{\mu}$ with $N_j(\bm{\mu};\alpha)$~\footnote{The notation $\bm{x}_j\,{=}\,\bm{\mu}$ means that the elements of the patch $\bm{x}_j$ encode the tuple of features $\bm{\mu}$.}. Normalizing this number by $N_j(\bm{\mu})\,{=}\,\sum_\alpha N_j(\bm{\mu};\alpha)$ yields the conditional probability $f_{j}(\alpha|\bm{\mu})$ for a datum to belong to class $\alpha$ conditioned on displaying the $s$-tuple $\bm{\mu}$ in the $j$-th input patch,
\begin{equation}
\label{eq:cond-freq-def2}
f_{j}(\alpha|\bm{\mu}) := \prob{\bm{x}\in\alpha|\bm{x}_{j}=\bm{\mu}} = \frac{N_{j}(\bm{\mu}; \alpha)}{N_{j}(\bm{\mu})}.
\end{equation}
If the low-level features are homogeneously spread across classes, then $f\,{=}\,n_c^{-1}$, independently of and $\alpha$, $\bm{\mu}$, and $j$. In contrast, due to the aforementioned correlations, the probabilities of the RHM are all different from $n_c^{-1}$---we refer to this difference as \emph{signal}. Distinct level-$1$ tuples $\bm{\mu}$ and $\bm{\nu}$ yield a different $f$ (and thus a different signal) with high probability unless $\bm{\mu}$ and $\bm{\nu}$ are synonyms, i.e. they share the same level-$2$ representation. Therefore, this signal can be used to identify synonymous level-$1$ tuples.

\subsection{Signal vs Sampling Noise} 
\begin{figure}
    \centering
    \includegraphics[width=.9\linewidth]{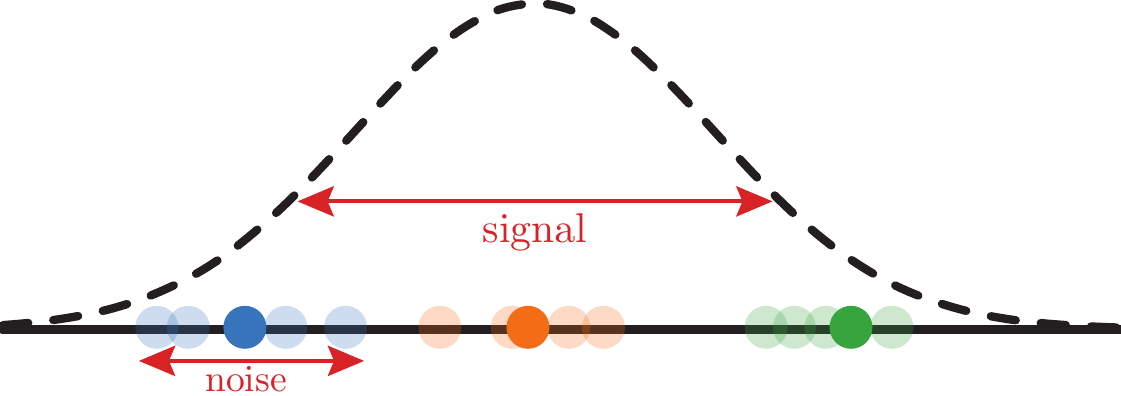}
    \caption{\textbf{Signal vs noise illustration.} The dashed function represents the distribution of $f(\alpha|\bm \mu)$ resulting from the random sampling of the RHM rules. The solid dots illustrate the \textit{true} frequencies $f(\alpha|\bm\mu)$ sampled from this distribution, with different colors corresponding to different groups of synonyms. The typical spacing between the solid dots, given by the width of the distribution, represents the \textit{signal}. Transparent dots represent the empirical frequencies $\hat{f}_{j}(\alpha|\bm\mu)$, with dots of the same color corresponding to synonymous features. The spread of transparent dots of the same color, which is due to the finiteness of the training set, represents the \textit{noise}.}
    \label{fig:signal-vs-noise}
\end{figure}

When measuring the conditional class probabilities with only $P$ training data, the occurrences in the right-hand side of~\autoref{eq:cond-freq-def2} are replaced with empirical occurrences, which induce a sampling \emph{noise} on the $f$'s. For the identification of synonyms to be possible, this noise must be smaller in magnitude than the aforementioned signal---a visual representation of the comparison between signal and noise is depicted in~\autoref{fig:signal-vs-noise}.

The magnitude of the signal can be computed as the ratio between the standard deviation and mean of $f_j(\alpha|\bm{\mu
})$ over realizations of the RHM. The full calculation is presented in~\autoref{app:correlations}: here we present a simplified argument based on an additional independence assumption. Given a class $\alpha$, the tuple $\bm{\mu}$ appearing in the $j$-th input patch is determined by a sequence of $L$ choices---one choice per level of the hierarchy---of one among $m$ possible lower-level representations. These $m^L$ possibilities lead to all the $mv$ distinct input $s$-tuples. $N_j(\bm{\mu};\alpha)$ is proportional to how often the tuple $\bm{\mu}$ is chosen---$m^L/(mv)$ times on average. Under the assumption of independence of the $m^L$ choices, the fluctuations of $N_{j}(\bm{\mu};\alpha)$ relative to its mean are given by the central limit theorem and read $(m^L/(m v))^{-1/2}$ in the limit of large $m$. If $n_c$ is sufficiently large, the fluctuations of $N_j(\bm{\mu})$ are negligible in comparison. Therefore, the relative fluctuations of $f_j$ are the same as those of $N_j(\bm{\mu};\alpha)$, and the size of the signal is $(m^L/(m v))^{-1/2}$.

The magnitude of the noise is given by the ratio between the standard deviation and mean, over independent samplings of a training set of fixed size $P$, of the empirical conditional probabilities $\hat{f}_{j}(\alpha|\bm{\mu})$. Only $P/(n_c m v)$ of the training points will, on average, belong to class $\alpha$ while displaying feature $\mu$ in the $j$-th patch. Therefore, by the convergence of the empirical measure to the true probability, the sampling fluctuations of $\hat{f}$ relative to the mean are of order $[P/(n_c m v)]^{-1/2}$---see~\autoref{app:correlations} for a detailed derivation. Balancing signal and noise yields the characteristic $P_c$ for the emergence of correlations. For large $m$, $n_c$ and $P$,
\begin{equation}\label{eq:sc-correlations}
P_c = n_c m^L,
\end{equation}
which coincides with the empirical sample complexity of deep networks discussed in~\autoref{sec:empirical}.

\subsection{Learning Level-1 Synonyms With One Step of Gradient Descent}\label{ssec:onestep}

To complete the argument, we consider a simplified one-step gradient descent setting~\cite{damian22neural, ba2022highdimensional}, where $P_c$ marks the number of training examples required to learn a synonymic invariant representation. In particular, we focus on the $s$-dimensional patches of the data and study how a two-layer network acting on one of such patches learns the first composition rule of the RHM by building a representation invariant to exchanges of level-$1$ synonyms.

Let us then sample an instance of the RHM, and $P$ input-label pairs $(\bm{x}_{k,1}, \alpha_k)$ with $\alpha_k\,{:=}\,\alpha(\bm{x}_k)$ for all $k\,{=}\,1,\dots,P$ and $\bm{x}_{k,1}$ denoting the first $s$-patch of the datum $\bm{x}_k$. The network output reads 
\begin{equation}\label{eq:two-layers-fcn}
\mathcal{F}_{\text{NN}}(\bm{x}_{1}) = \frac{1}{H}\sum_{h=1}^H a_h \sigma(\bm{w}_h \cdot \bm{x}_{1}),
\end{equation}
where the inner-layer weights $\bm{w}_h$'s have the same dimension as $\bm{x}_{1}$, the top-layer weights $a_h$'s are $n_c$-dimensional and $\sigma(x)\,{=}\,\text{max}\left(0,x\right)$ is the ReLU activation function. To further simplify the problem, we represent $\bm{x}_1$ as a $v^s$-dimensional one-hot encoding of the corresponding $s$-tuple of features. This representation is equivalent to an orthogonalization of the input points. In addition, the top-layer weights are initialized as i.i.d. Gaussian with zero mean and unit variance and fixed, whereas the $\bm{w}_h$'s are initialized with all their elements set to $1$ and trained by Gradient Descent (GD) on the empirical cross-entropy loss,
\begin{align}\label{eq:cross-ent-loss}
\mathcal{L} = \frac{1}{P}\displaystyle\sum_{k=1}^P \left[ -\log{\left(\frac{e^{\left(\mathcal{F}_{\text{NN}}(\bm{x}_{k,1})\right)_{\alpha(\bm{x}_k)}}}{\sum_{\beta=1}^{n_c} e^{\left(\mathcal{F}_{\text{NN}}(\bm{x}_{k,1})\right)_{\beta}}}\right)} \right].
\end{align}
Finally, we consider the mean-field limit $W\to\infty$, so that, at initialization, $\mathcal{F}_{\text{NN}}^{(0)}\,{=}\,0$ identically.

Let us denote with $\bm{\mu}(\bm{x}_1)$ the $s$-tuple of features encoded in $\bm{x}_1$. Due to the one-hot encoding, $f_h(\bm{x}_1)\,{:=}\,\bm{w}_h\cdot \bm{x}_1$ coincides with the $\bm{\mu}(\bm{x}_1)$-th component of the weight $\bm{w}_h$. This component, which is set to $1$ at initialization, is updated by (minus) the corresponding component of the gradient of the loss in~\autoref{eq:cross-ent-loss}. Recalling also that the predictor is $0$ at initialization, we get
\begin{align}\label{eq:rep-gradient-proof}
&\Delta f_h(\bm{x}_1) = -\nabla_{\left(\bm{w}_h\right)_{\bm{\mu}(\bm{x}_1)}} \mathcal{L} = \nonumber\\
& \frac{1}{P}\sum_{k=1}^P \sum_{\alpha=1}^{n_c}a_{h,\alpha} \delta_{\mu(\bm{x}_1), \mu(\bm{x}_{k,1})} \left(\delta_{\alpha,\alpha(\bm{x}_k)}-\frac{1}{n_c}\right)  =\nonumber\\
&\sum_{\alpha=1}^{n_c} a_{h,\alpha} \left(\frac{\hat{N}_{1}(\bm{\mu}(\bm{x}_1);\alpha)}{P} - \frac{1}{n_c}\frac{\hat{N}_{1}(\bm{\mu})}{P} \right),
\end{align}
where $\hat{N}_1(\bm{\mu})$ is the empirical occurrence of the $s$-tuple $\bm{\mu}$ in the first patch of the $P$ training points and $\hat{N}_1(\bm{\mu};\alpha)$ is the (empirical) joint occurrence of the $s$-tuple $\bm{\mu}$ and the class label $\alpha$. As $P$ increases, the empirical occurrences $\hat{N}$ converge to the true occurrences $N$, which are invariant for the exchange of synonym $s$-tuples $\bm{\mu}$. Hence, the hidden representation is also invariant for the exchange of synonym $s$-tuples in this limit.

\begin{figure}[h]
    \centering
    \includegraphics[width=\linewidth]{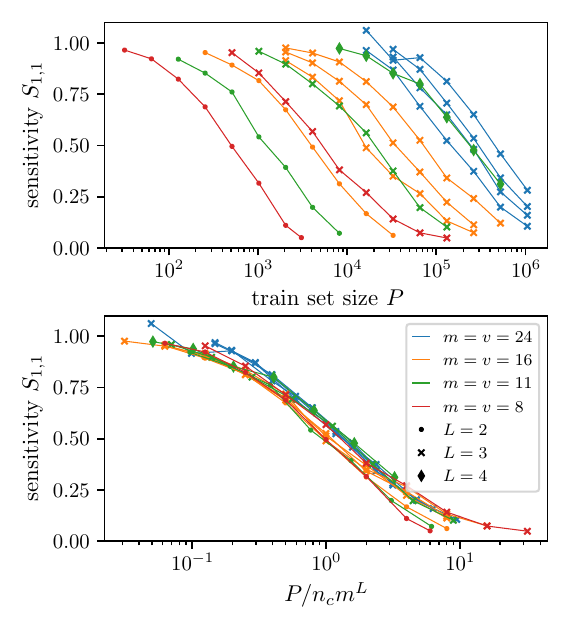}
    \caption{\textbf{Synonymic sensitivity of the hidden representation vs $P$ for a two-layer fully-connected network} trained on the first patch of the inputs of an RHM with $s\,{=}\,2$ and $m\,{=}\,v$, for varying $L$, $v$, and $n_c$. The top panel shows the bare curves whereas, in the bottom panel, the x-axis is rescaled by $P_c = n_c m^{L}$. The collapse of the rescaled curves highlights that $P_c$ coincides with the number of training data for building a synonymic invariant representation.
    }
    \label{fig:ortho-noisy}
\end{figure}
This prediction is confirmed empirically in~\autoref{fig:ortho-noisy}, which shows the sensitivity $S_{1,1}$ of the hidden representation~\footnote{Here invariance to exchange of level-$1$ synonyms can already be achieved at the first hidden layer due to the orthogonalization of the $s$-dimensional patches of the input, which makes them linearly separable.} of shallow fully-connected networks trained in the setting of this section, as a function of the number $P$ of training data for different combinations of the model parameters. The bottom panel, in particular, highlights that the sensitivity is close to $1$ for $P\,{\ll}\,P_c$ and close to $0$ for $P\,{\gg}\,P_c$. In addition, notice that the collapse of the pre-activations of synonymic tuples onto the same, synonymic invariant value, implies that the rank of the hidden weights matrix tends to $v$---the vocabulary size of higher-level features. This low-rank structure is typical in the weights of deep networks trained on image classification~\cite{denil2013predicting,denton2014exploiting,yu2017compressing,guth2023rainbow}.

\paragraph{Including all patches via weight sharing.} Let us remark that one can easily extend the one-step setting to include the information from all the input patches, for instance by replacing the network in~\autoref{eq:two-layers-fcn} with a one-hidden-layer convolutional network with filter size $s$ and nonoverlapping patches. Consequently, the empirical occurrences on the right-hand side of~\autoref{eq:rep-gradient-proof} would be replaced with average occurrences over the patches. However, this average results in a reduction of both the signal and the sampling noise contributions to the empirical occurrences by the same factor $\sqrt{s^{L-1}}$. Therefore, weight sharing does not affect the sample size required for synonymic invariance in the one-step setting.

\paragraph{Improved sample complexity via clustering.} A distance-based clustering method acting on the representations of~\autoref{eq:rep-gradient-proof} can actually identify synonyms at $P\,{\simeq}\,\sqrt{n_c}m^L\,{=}\,P_c/\sqrt{n_c}$, which is much smaller than $P_c$ in the large-$n_c$ limit. Intuitively, using a sequence instead of a scalar amplifies the signal by a factor $n_c$ and the sampling noise by a factor $\sqrt{n_c}$, improving the signal-to-noise ratio. We show that this is indeed the case in~\autoref{app:clustering} for the maximal dataset case $n_c\,{=}\,v$ and $m\,{=}\,v^{s-1}$. Previous theoretical studies have considered the possibility of intercalating clustering steps in standard gradient descent methods~\cite{malach2018provably, malach2020implications}, but the question of whether deep learning methods can achieve a similar sample complexity with standard end-to-end training remains open.

\subsection{Curse of Dimensionality without Correlations}\label{cod_nocorr}

\begin{figure}[h]
    \centering
    \includegraphics[width=.45\textwidth]{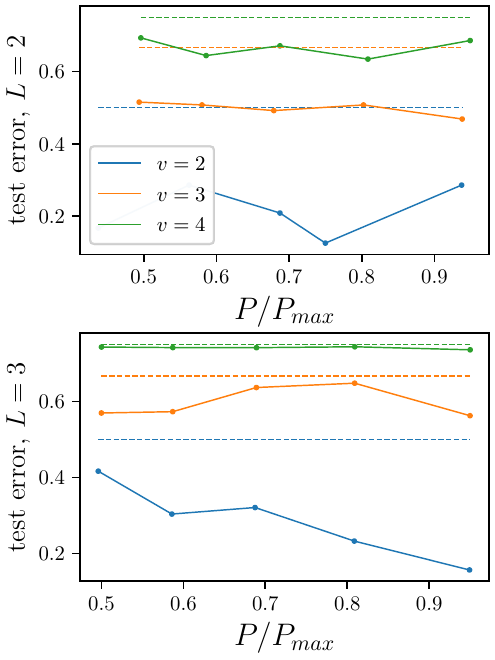}
    \caption{\textbf{Test error of depth-($L+1$) CNNs trained on uncorrelated RHM vs number $P$ of training points} rescaled with $P_\text{max}$, with $s\,{=}\,2$ and $m\,{=}\,n_c\,{=}\,v$ with different $v$ (different colors), for $L=2$ (top) and $L=3$ (bottom). Horizontal dashed lines stand for $\epsilon_{\text{rand}}$, given by guessing the label uniformly at random.}
    \label{fig:parity_main}
\end{figure}

To support the argument that learning is possible because of the detection of local input-label correlations, we show that their removal in the RHM leads to a sample complexity exponential in $d$, even for deep networks. Removing such correlations implies that, at any level, features are uniformly distributed among classes. This is achieved enforcing that a tuple $\bm{\mu}$ in the $j-$th patch at level $\ell$ belongs to a class $\alpha$ with probability $n_c^{-1}$, independently on $\bm{\mu}$, $j$, $\ell$ and $\alpha$, as discussed in \autoref{subsec:counting}. Such procedure produces an uncorrelated version of the RHM, which generalizes the parity problem (realized for $m=v=n_c=2$), a task that cannot be learned efficiently with gradient-based methods~\cite{shalev2017failures}. Indeed, deep CNNs with depth $L+1$, trained on this uncorrelated RHM, are cursed by dimensionality, as shown in \autoref{fig:parity_main}. The CNN test error is close to $\epsilon_{\text{rand}}$, given by randomly guessing the label, even for $P/P_{\text{max}}>0.9$, particularly for $v>2$.

\section{Conclusion}

What makes real-world tasks learnable? This question extends from machine learning to brain science~\cite{kruger2012deep}. To start thinking quantitatively about it, we introduced the Random Hierarchy Model: a family of tasks that captures the compositional structure of natural data. We showed that neural networks can learn such tasks with a limited training set, by developing a hierarchical representation of the data. Overall, these results rationalize several phenomena associated with deep learning.

First, our finding that for hierarchical tasks, the sample complexity is polynomial in the input dimension (and not exponential) leads to a plausible explanation for the learnability of real-world tasks. Moreover, our results provide a rule of thumb for estimating the order of magnitude of the sample complexity of benchmark datasets. In the case of CIFAR10 \cite{krizhevsky_learning_2009}, for instance, having $10$ classes, taking reasonable values for task parameters such as $m\in[5,15]$ and $L=3$, yields $P^* \in [10^3, 3\times 10^4]$, comparable with the sample complexity of modern architectures (see~\autoref{fig:terr_vs_P_cifar10}).

Secondly, our results quantify the intuition that depth is crucial to building a hierarchical representation that effectively lowers the dimension of the problem, and allows for avoiding the curse of dimensionality. On the one hand, this result gives a  foundation to the claim that deep is better than shallow, beyond previous analyses that focused on expressivity~\cite{poggio2017why, schmidt2020nonparametric} rather than learning.
On the other hand, our result that the internal representations of trained networks mirror the hierarchical structure of the task explains why these representations become increasingly complex with depth in real-world applications ~\cite{zeiler_visualizing_2014,doimo2020hierarchical}.

Furthermore, we provided a characterization of the internal representations based on their sensitivity towards transformations of the low-level features that leave the class label unchanged. This viewpoint complements existing ones that focus instead on the input features that maximize the response of hidden neurons, thus enhancing the interpretability of neural nets. In addition, our approach bypasses several issues of previous characterizations. For example, approaches based on mutual information~\cite{shwartz2017opening} are ill-defined when the network representations are deterministic functions of the input~\cite{saxe2019information}, whereas those based on intrinsic dimension~\cite{ansuini2019intrinsic, recanatesi2019dimensionality} can display counterintuitive results---see~\autoref{app:intrinsic-dimension} for a deeper discussion of the intrinsic dimension and on how it behaves in our framework.

Finally, our study predicts a fundamental relationship between sample complexity, correlations between low-level features and labels, and the emergence of invariant representations. This prediction can be tested beyond the context of our model, for instance by studying invariance to exchanging synonyms in language modeling tasks.

Looking forward, the Random Hierarchy Model is a suitable candidate for the clarification of other open questions in the theory of deep learning. For instance, a formidable challenge is to obtain a detailed description of the gradient-descent dynamics of deep networks. Indeed, dynamics may be significantly easier to analyze in this model, since quantities characterizing the network success, such as sensitivity to synonyms, can be delineated. In addition, the model could be generalized to describe additional properties of data, e.g., noise in the form of errors in the composition rules or inhomogeneities in the frequencies at which high-level features generate low-level representations. The latter, in particular, would generate data where certain input features are more abundant than others and, possibly, to a richer learning scenario with several characteristic training set sizes.

Beyond supervised learning, in the Random Hierarchy Model the set of available input data inherits the hierarchical structure of the generative process. Thus, this model offers a new way to study the effect of compositionality on self-supervised learning or probabilistic generative models---extremely powerful techniques whose understanding is still in its infancy.

\section*{Acknowledgements}

The authors thank Antonio Sclocchi for fruitful discussions and helpful feedback on the manuscript. This work was supported by a grant from the Simons Foundation (\#454953
Matthieu Wyart).

\bibliography{main}

\appendix

\section{Methods}\label{sec:matmet}

\subsection{RHM implementation}
The code implementing the RHM is available online at \url{https://github.com/pcsl-epfl/hierarchy-learning/blob/master/datasets/hierarchical.py}.
The inputs sampled from the RHM are represented as a one-hot encoding of low-level features so that each input consists of $s^L$ pixels and $v$ channels (size $s^L \times v$). The input pixels are whitened over channels, i.e., each pixel has zero mean and unit variance over the channels.

\subsection{Machine Learning Models}

We consider both generic deep neural networks and deep convolutional networks (CNNs) tailored to the structure of the RHM. Generic deep neural networks are made by stacking \emph{fully-connected} layers, i.e., linear transformations of the kind
\begin{equation}\label{eq:linear}
x\in \mathbb{R}^{d_{\text{in}}} \to d_{\text{in}}^{-1/2} W\cdot x + b \in \mathbb{R}^{d_{\text{out}}},
\end{equation}
where $W$ is a $d_{\text{out}}\times d_{\text{in}}$ matrix of weights, $b$ a $d_{\text{out}}$ sequence of biases, and the factor $d_{\text{in}}^{-1/2}$ guarantees that the outputs remain of order $1$ when $d_{\text{in}}$ is varied. \emph{Convolutional} layers, instead, act on image-like inputs that have a spatial dimension $d$ and $c_{\text{in}}$ channels and compute the convolution of the input with a filter of spatial size $f$. This operation is equivalent to applying the linear transformation of~\autoref{eq:linear} to input patches of spatial size $f$, i.e., groups of $f$ adjacent pixels (dimension $d_{\text{in}}\,{=}\, (f \times c_{\text{in}})$). The output has an image-like structure analogous to that of the input, with spatial dimension depending on how many patches are considered. In the \emph{nonoverlapping patches} case, for instance, the spatial dimension of the output is $d/f$.

For all layers but the last, the linear transformation is followed by an element-wise nonlinear activation function $\sigma$. We resort to the popular Rectified Linear Unit (ReLU) $\sigma(x)\,{=}\,\text{max}\left(0,x\right)$. The output dimension is always fixed to the number of classes $n_c$, while the input dimension of the first layer is the same as the input data: spatial dimension $s^L$ and $v$ channels, flattened into a single $s^L\times v$ sequence when using a fully-connected layer. The dimensionalities of the other \emph{hidden} layers are set to the same constant $H$ throughout the network. Following the maximal update parametrization~\cite{yang2019scaling}, the weights of the last layer are multiplied by an additional factor ${H}^{-1}$. This factor causes the output at initialization to vanish as $H$ grows, which induces representation learning even in the $H\to\infty$ limit. In practice, we set $H\,{=}\,\left(4- 8\right)\times v^{s}$. Increasing this number further does not affect any of the results presented in the paper.

To tailor deep CNNs to the structure of the RHM, we set $f\,{=}\,s$ so that, in the nonoverlapping patches setting, each convolutional filter acts on a group of $s$ low-level features that correspond to the same higher-level feature. Since the spatial dimensionality of the input is $s^L$ and each layer reduces it by $s$, the number of nonlinear layers in a tailored CNN is fixed to the depth of the RHM $L$, so that the network depth is $L\,{+}\,1$. Fully-connected networks, instead, can have any depth. The code for the implementation of both architectures is available at~\url{https://github.com/pcsl-epfl/hierarchy-learning/blob/master/models}.

\subsection{Training Procedure}

Training is performed within the PyTorch deep learning framework \cite{paszke_pytorch_2019}. Neural networks are trained on $P$ training points sampled uniformly at random from the RHM data, using stochastic gradient descent (SGD) on the cross-entropy loss. The batch size is $128$ for $P\,{\geq}\,128$ and $P$ otherwise, the learning rate is initialised to $10^{-1}$ and follows a cosine annealing schedule which reduces it to $10^{-2}$ over $100$ epochs. Training stops when the training loss reaches $10^{-3}$. The corresponding code is available at~\url{https://github.com/pcsl-epfl/hierarchy-learning/blob/master}

The performance of the trained models is measured as the classification error on a test set. The size of the test set is set to $\min(P_{\max} - P,\, 20'000)$. Synonymic sensitivity, as defined in \autoref{eq:stability-def}, is measured on a test set of size $\min(P_{\max} - P,\, 1'000)$. Reported results for a given value of RHM parameters are averaged over 10 jointly different instances of the RHM and network initialization.

\section{Statistics of The Composition Rules}\label{app:single-rule}

In this section, we consider a single composition rule, that is the assignment of $m$ $s$-tuples of low-level features to each of the $v$ high-level features. In the RHM these rules are chosen uniformly at random over all the possible rules, thus their statistics are crucial in determining the correlations between the input features and the class label.

\subsection{Statistics of a single rule}

For each rule, we call $N_i(\mu_1;\mu_2)$ the number of occurrences of the low-level feature $\mu_1$ in position $i$ of the $s$-tuples generated by the higher-level feature $\mu_2$. The probability of $N_i(\mu_1;\mu_2)$ is that of the number of successes when drawing $m$ (number of $s$-tuples associated with the high-level feature $\mu_2$) times without replacement from a pool of $v^s$ (total number of $s$-tuples with vocabulary size $v$) objects where only $v^{s-1}$ satisfy a certain condition (number of $s$-tuples displaying feature $\mu_1$ in position $i$):
\be
\prob{ N_i(\mu_0;\mu_1) = k} = \binom{v^{s-1}}{k}\binom{v^s-v^{s-1}}{m-k}\left/\binom{v^s}{m}\right.,
\ee
which is a Hypergeometric distribution $\Hg{v^s, v^{s-1}, m}$, with mean
\be
\avg{N} = m\frac{v^{s-1}}{v^s} = \frac{m}{v},
\ee
and variance
\begin{align}
\sigma_N^2 &:= \avg{\left(N-\avg{N}\right)^2} = m\frac{v^{s-1}}{v^s}\frac{v^s-v^{s-1}}{v^s}\frac{v^s-m}{v^s-1}\nonumber\\
&= \frac{m}{v}\frac{v-1}{v} \frac{v^s-m}{v^s-1} \xrightarrow{m\,{\gg}\,1} \frac{m}{v},
\end{align}
independently of the position $i$ and the specific low- and high-level features. Notice that, since $m\,{\leq}\, v^{s-1}$ with $s$ fixed, large $m$ implies also large $v$.

\subsection{Joint statistics of a single rule} 

\paragraph{Shared high-level feature.} For a fixed high-level feature $\mu_2$, the joint probability of the occurrences of two different low-level features $\mu_1$ and $\nu_1$ is a multivariate Hypergeometric distribution,
\begin{align}\label{eq:hypergeo-multi}
&\prob{ N_i(\mu_1;\mu_2) = k; N_i(\nu_1;\mu_2) = l} \nonumber\\
&= \binom{v^{s-1}}{k}\binom{v^{s-1}}{l}\binom{v^s-2v^{s-1}}{m-k-l}\left/\binom{v^s}{m}\right.,
\end{align}
giving the following covariance,
\begin{align}\label{eq:cov-interclass}
c_{N}  :=&  \avg{\left(N_i(\mu_1;\mu_2)-\avg{N}\right)\left(N_i(\nu_1;\mu_2)-\avg{N}\right)}\nonumber\\ =& -\frac{m}{v^2}\frac{v^s-m}{v^s-1} \xrightarrow{m\,{\gg}\,1} -\left(\frac{m}{v}\right)^2\frac{1}{m}.
\end{align}
The covariance can also be obtained via the constraint $\sum_{\mu_1} N_i(\mu_1;\mu_2)\,{=}\,m$. For any finite sequence of identically distributed random variables $X_\mu$ with a constraint on the sum $\sum_\mu X_\mu \,{=}\, m$,
\begin{align}\label{eq:var-cov}
&\sum_{\mu=1}^v X_\mu \,{=}\, m \Rightarrow \sum_{\mu=1}^v (X_\mu -\avg{X_\mu}) = 0 \Rightarrow\nonumber\\
& (X_\nu -\avg{X_\nu})\sum_{\mu=1}^v (X_\mu -\avg{X_\mu}) = 0 \Rightarrow\nonumber\\
& \sum_{\mu=1}^v \avg{(X_\nu -\avg{X_\nu})(X_\mu -\avg{X_\mu})} = 0 \Rightarrow\nonumber\\
& \var{X_\mu} + (v-1) \cov{X_\mu, X_\nu} = 0.
\end{align}
In the last line, we used the identically distributed variables hypothesis to replace the sum over $\mu\neq \nu$ with the factor $(v-1)$. Therefore,
\begin{align}
c_N &= \cov{N_i(\mu_1;\mu_2), N_i(\nu_1;\mu_2)} \nonumber\\ &= -\frac{\var{N_i(\mu_1;\mu_2)}}{v-1} = -\frac{\sigma^2_N}{v-1}.
\end{align}

\paragraph{Shared low-level feature.} The joint probability of the occurrences of the same low-level feature $\mu_1$ starting from different high-level features $\mu_2\,{\neq}\,\nu_2$ can be written as follows,
\begin{align}
&\prob{ N(\mu_1;\mu_2) = k; N(\mu_1;\nu_2) = l} = \nonumber\\
&\prob{N(\mu_1;\mu_2) = k|N(\mu_1;\nu_2) = l} \times \prob{N(\mu_1;\nu_2) = l} =\nonumber\\
& \Hg{v^s-m,v^{s-1}-l,m}(k) \times \Hg{v^s,v^{s-1},m}(l),
\end{align}
resulting in the following `inter-feature' covariance,
\begin{align}\label{eq:cov-interfeat}
c_{if} := \cov{N_i(\mu_1;\mu_2),N_i(\mu_1;\nu_2)} =  -\left(\frac{m}{v}\right)^2\frac{v-1}{v^s-1}.
\end{align}

\paragraph{No shared features.} Finally, by multiplying both sides of $\sum_{\mu_1} N(\mu_1;\mu_2)\,{=}\,m$ with $N(\nu_1;\nu_2)$ and averaging, we get
\begin{align}\label{eq:cov-general}
&c_{g} := \cov{N_i(\mu_1;\mu_2),N_i(\nu_1;\nu_2)} = \nonumber\\
&-\frac{\cov{N_i(\mu_1;\mu_2),N_i(\mu_1;\nu_2)}}{v-1} =  \left(\frac{m}{v}\right)^2 \frac{1}{v^s-1}.
\end{align}

\section{Emergence of input-output correlations ($P_c$)}\label{app:correlations}

As discussed in the main text, the Random Hierarchy Model presents a characteristic sample size $P_c$ corresponding to the emergence of the input-output correlations. This sample size predicts the sample complexity of deep CNNs, as we also discuss in the main text. In this appendix, we prove that
\be
P_c \xrightarrow{n_c, m\to \infty} n_c m^L.
\ee

\subsection{Estimating the Signal}

The correlations between input features and the class label can be quantified via the conditional probability (over realizations of the RHM) of a data point belonging to class $\alpha$ conditioned on displaying the $s$-tuple $\bm{\mu}$ in the $j$-th input patch,
\be\label{eq:cond-freq-def}
f_{j}(\alpha|\bm{\mu}) := \prob{\bm{x}\in\alpha|\bm{x}_{j}=\bm{\mu}},
\ee
where the notation $\bm{x}_j\,{=}\,\bm{\mu}$ means that the elements of the patch $\bm{x}_j$ encode the tuple of features $\bm{\mu}$. We say that the low-level features are correlated with the output if
\be
f_j(\alpha|\bm{\mu}) \neq \frac{1}{n_c},
\ee
and define a `signal' as the difference $f_j(\alpha|\bm{\mu})\,{-}\,n_c^{-1}$. In the following, we compute the statistics of the signal over realizations of the RHM. 

\subsubsection{Occurrence of low-level features}

Let us begin by defining the joint occurrences of a class label $\alpha$ and a low-level feature $\mu_1$ in a given position of the input. Using the tree representation of the model, we will identify an input position with a set of $L$ indices $i_{\ell}\,{=}\,1,\dots,s$, each indicating which branch to follow when descending from the root (class label) to a given leaf (low-level feature). These joint occurrences can be computed by combining the occurrences of the single rules introduced in~\autoref{app:single-rule}. With $L\,{=}\,2$, for instance,
\be\label{eq:occ-l2}
N^{(1\to 2)}_{i_1 i_2}(\mu_1;\alpha) = \sum_{\mu_2=1}^v \left(m^{s-1} N^{(1)}_{i_1}(\mu_1;\mu_2) \right)\times  N^{(2)}_{i_2}(\mu_2;\alpha),
\ee
where:
\begin{itemize}
\item[\emph{i)}] $N^{(2)}_{i_2}(\mu_2;\alpha)$ counts the occurrences of $\mu_2$ in position $i_2$ of the level-$2$ representations of $\alpha$, i.e. the $s$-tuples generated from $\alpha$ according to the second-layer composition rule;
\item[\emph{ii)}] $N^{(1)}_{i_1}(\mu_1;\mu_2)$ counts the occurrences of $\mu_1$ in position $i_1$ of the level-$1$ representations of $\mu_2$, i.e. $s$-tuples generated by $\mu_2$ according to the composition rule of the first layer;
\item[\emph{iii)}] the factor $m^{s-1}$ counts the descendants of the remaining $s-1$ elements of the level-$2$ representation ($m$ descendants per element); 
\item[\emph{iv)}] the sum over $\mu_2$ counts all the possible paths of features that lead to $\mu_1$ from $\alpha$ across $2$ generations.
\end{itemize}
The generalization of~\autoref{eq:occ-l2} is immediate once one takes into account that the multiplicity factor accounting for the descendants of the remaining positions at the $\ell$-th generation is equal to $m^{s^{\ell-1}}/m$ ($s^{\ell-1}$ is the size of the representation at the previous level). Hence, the overall multiplicity factor after $L$ generations is
\be
1\times \frac{m^{s}}{m} \times \frac{m^{s^2}}{m} \times \dots \times \frac{m^{s^{L-1}}}{m} = m^{\frac{s^L-1}{s-1}-L},
\ee
so that the number of occurrences of feature $\mu_1$ in position $i_1\dots i_L$ of the inputs belonging to class $\alpha$ is
\begin{widetext}
\be\label{eq:occ-single}
N^{(1\to L)}_{i_{1\to L}}(\mu_1;\alpha)  = m^{\frac{s^L-1}{s-1}-L} \sum_{\mu_2, \dots, \mu_{L}=1}^v N^{(1)}_{i_1}(\mu_1;\mu_2) \times\dots\times  N^{(L)}_{i_L}(\mu_{L};\alpha),
\ee
\end{widetext}
where we used $i_{1\to L}$ as a shorthand notation for the tuple of indices $i_1, i_2, \dots,i_L$. 

The same construction allows us to compute the number of occurrences of up to $s-1$ features within the $s$-dimensional patch of the input corresponding to the path $i_{2\to L}$. The number of occurrences of a whole $s$-tuple, instead, follows a slightly different rule, since there is only one level-$2$ feature $\mu_2$ which generates the whole $s$-tuple of level-$1$ features $\bm{\mu}_1\,{=}\,(\mu_{1,1},\dots,\mu_{1,s})$---we call this feature $g_1(\bm{\mu}_1)$, with $g_1$ denoting the first-layer composition rule. As a result, the sum over $\mu_2$ in the right-hand side of~\autoref{eq:occ-single} disappears and we are left with
\begin{widetext}
\be\label{eq:occ-tuple}
N^{(1\to L)}_{i_{2\to L}}(\bm{\mu}_1;\alpha)  = m^{\frac{s^L-1}{s-1}-L} \sum_{\mu_{3}, \dots, \mu_{L}=1}^v N^{(2)}_{i_2}(g_1(\bm{\mu}_1);\mu_3) \times\dots\times  N^{(L)}_{i_L}(\mu_{L};\alpha).
\ee
\end{widetext}
Coincidentally,~\autoref{eq:occ-tuple} shows that the joint occurrences of a $s$-tuple of low-level features $\bm{\mu}_1$ depend on the level-$2$ feature corresponding to $\bm{\mu}_1$. Hence, $N^{(1\to L)}_{i_{2\to L}}(\bm{\mu}_1;\alpha)$ is invariant for the exchange of $\bm{\mu}_1$ with one of its synonyms, i.e. level-$1$ tuples $\bm{\nu}_1$ corresponding to the same level-$2$ feature.

\begin{widetext}

\subsubsection{Class probability conditioned on low-level observations}

We can turn these numbers into probabilities by normalizing them appropriately. Upon dividing by the total occurrences of a low-level feature $\mu_1$ independently of the class, for instance, we obtain the conditional probability of the class of a given input, conditioned on the feature in position $i_1\dots i_L$ being $\mu_1$.
\be\label{eq:cond-freq}
f^{(1\to L)}_{i_{1\to L}}(\alpha|\mu_1) := \frac{N^{(1\to L)}_{i_{1\to L}}(\mu_1;\alpha)}{\displaystyle\sum_{\alpha'=1}^{n_c} N^{(1 \to L)}_{i_{1\to L}}(\mu_1;\alpha')} = \frac{\displaystyle\sum_{\mu_2, \dots, \mu_{L}=1}^v N^{(1)}_{i_1}(\mu_1;\mu_2) \times\dots\times  N^{(L)}_{i_L}(\mu_{L};\alpha)}{\displaystyle\sum_{\mu_2, \dots, \mu_{L}=1}^v\sum_{\mu_{L+1}=1}^{n_c} N^{(1)}_{i_1}(\mu_1;\mu_2) \times\dots\times  N^{(L)}_{i_L}(\mu_{L};\mu_{L+1})}.
\ee
Let us also introduce, for convenience, the numerator and denominator of the right-hand side of~\autoref{eq:cond-freq}.
\be
U^{(1\to L)}_{i_{1\to L}}(\mu_1\alpha) = \displaystyle\sum_{\mu_2, \dots, \mu_{L}=1}^v N^{(1)}_{i_1}(\mu_1;\mu_2) \times\dots\times  N^{(L)}_{i_L}(\mu_{L};\alpha);\quad D^{(1\to L)}_{i_{1\to L}}(\mu_1) = \displaystyle\sum_{\alpha=1}^{n_c} U^{(1\to L)}_{i_{1\to L}}(\mu_1;\alpha).
\ee

\subsubsection{Statistics of the numerator $U$}

We now determine the first and second moments of the numerator of $f^{(1\to L)}_{i_{1\to L}}(\mu_1;\alpha)$. Let us first recall the definition for clarity,
\be\label{eq:num-def}
U^{(1\to L)}_{i_{1\to L}}(\mu_1;\alpha) = \displaystyle\sum_{\mu_2, \dots, \mu_{L}=1}^v N^{(1)}_{i_1}(\mu_1;\mu_2) \times\dots\times  N^{(L)}_{i_L}(\mu_{L};\alpha)
\ee

\paragraph{Level 1 $L\,{=}\,1$.} For $L\,{=}\,1$, $U$ is simply the occurrence of a single production rule $N_i(\mu_1;\alpha)$,
\begin{align}
\avg{U^{(1)}} &= \frac{m}{v};\\
\sigma^2_{U^{(1)}}:=\var{U^{(1)}} &= \frac{m}{v}\frac{v-1}{v}\frac{v^s-m}{v^s-1} \xrightarrow{v\gg 1} \frac{m}{v};\\
c_{U^{(1)}}:=\cov{U^{(1)}(\mu_1;\alpha), U^{(1)}(\nu_1;\alpha)} &= -\frac{\var{U^{(1)}}}{(v-1)} =  -\left(\frac{m}{v}\right)^2\frac{v^s-m}{v^s-1}\frac{1}{m} \xrightarrow{v\gg 1}\left(\frac{m}{v}\right)^2\frac{1}{m};
\end{align}
where the relationship between variance and covariance is due to the constraint on the sum of $U^{(1)}$ over $\mu_1$, see~\autoref{eq:var-cov}.

\paragraph{Level 2 $L\,{=}\,2$.} For $L\,{=}\,2$,
\be
U^{(1\to 2)}_{i_{1\to 2}}(\mu_1;\alpha) = \displaystyle\sum_{\mu_2=1}^v N^{(1)}_{i_1}(\mu_1;\mu_2) \times N^{(2)}_{i_2}(\mu_2;\alpha) = \sum_{\mu_2=1}^v N^{(1)}_{i_1}(\mu_1;\mu_2) U^{(2)}_{i_2}(\mu_2;\alpha).
\ee
Therefore,
\begin{align}
\avg{U^{(1\to 2)}} &= v\left(\frac{m}{v}\right) \times \avg{U^{(1)}} =  v\left(\frac{m}{v}\right)^2;\\
\sigma^2_{U^{(2)}}:=\var{U^{(1\to 2)}} &= \sum_{\mu_2,\nu_2=1}^v \left(\avg{N^{(1)}(\mu_1;\mu_2) N^{(1)}(\mu_1;\nu_2)}\avg{U^{(2)}(\mu_2;\alpha)U^{(2)}(\nu_2;\alpha)}-\avg{N}^2\avg{U^{(1)}}^2\right)\nonumber \\
&= \sum_{\mu_2, \nu_2=\mu_2} \dots + \sum_{\mu_2}\sum_{\nu_2 \neq \mu_2} \dots \nonumber \\
& = v\left(\sigma^2_N \sigma^2_{U^{(1)}} + \sigma^2_N \avg{U^{(1)}}^2 + \sigma^2_{U^{(1)}} \avg{N}^2  \right) + v(v-1)\left( c_{if} c_{U^{(1)}} + c_{if} \avg{U^{(1)}}^2 + c_{U^{(1)}} \avg{N}^2\right) \nonumber \\
& = v\left( \sigma^2_N \sigma^2_{U^{(1)}} +(v-1)c_{if} c_{U^{(1)}} \right) + v\avg{U^{(1)}}^2\left(\sigma_N^2 + (v-1)c_{if}\right) + v\avg{N}^2\left(\sigma_{U^{(1)}}^2 + (v-1)c_{U^{(1)}}\right) \nonumber \\
&= v \sigma^2_{U^{(1)}} \left( \sigma^2_N - c_{if} \right) + v\avg{U^{(1)}}^2\left(\sigma_N^2 + (v-1)c_{if}\right), \\
c_{U^{(2)}} &= -\frac{\sigma^2_{U^{(2)}}}{(v-1)}
\end{align}

\paragraph{Level L.} In general,
\be
U^{(1\to L)}_{i_{1\to L}}(\mu_1;\alpha) = \displaystyle\sum_{\mu_2=1}^v N^{(1)}_{i_1}(\mu_1;\mu_2) U^{(2\to L)}_{i_{2\to L}}(\mu_2;\alpha).
\ee
Therefore,
\begin{align}\label{eq:num-stat-L}
\avg{U^{(L)}} &= v\left(\frac{m}{v}\right)\times \avg{U^{(L-1)}} = v^{L-1}\left(\frac{m}{v}\right)^L;\\
\sigma^2_{U^{(L)}} &= \sum_{\mu_2,\nu_1=1}^v \left(\avg{N^{(1)}(\mu_1;\mu_2) N^{(1)}(\mu_1;\nu_2)}\avg{U^{(2\to L)}(\mu_2;\alpha)U^{(2\to L)}(\nu_1;\alpha)}-\avg{N}^2\avg{U^{(1\to (L-1))}}^2\right)\nonumber \\
&= \sum_{\mu_2, \nu_2=\mu_2} \dots + \sum_{\mu_2}\sum_{\nu_2 \neq \mu_2} \dots \nonumber \\
& = v\left(\sigma^2_N \sigma^2_{U^{(L-1)}} + \sigma^2_N \avg{U^{(L-1)}}^2 + \sigma^2_{U^{(L-1)}} \avg{N}^2  \right) + v(v-1)\left( \sigma^2_{if} c_{U^{(L-1)}} + c_{if} \avg{U^{(L-1)}}^2 + c_{U^{(L-1)}} \avg{N}^2\right) \nonumber \\
&= v \sigma^2_{U^{(L-1)}} \left( \sigma^2_N - c_{if} \right) + v\avg{U^{(L-1)}}^2\left(\sigma_N^2 + (v-1)c_{if}\right), \\
c_{U^{(L)}} &= -\frac{\sigma^2_{U^{(L)}}}{(v-1)}
\end{align}

\paragraph{Concentration for large $m$.} In the large multiplicity limit $m\,{\gg}\,1$, the $U$'s concentrate around their mean value. Due to $m\,{\leq}\,v^{s-1}$, large $m$ implies large $v$, thus we can proceed by setting $m\,{=}\, q v^{s-1}$, with $q\in(0,1]$ and studying the $v\,{\gg}\,1$ limit. From~\autoref{eq:num-stat-L},
\be
\avg{U^{(L)}} = q^L v^{L(s-1)-1}.
\ee
In addition,
\be
\sigma^2_N \xrightarrow{v\gg 1} \frac{m}{v} = q v^{(s-1)-1},\quad c_{if} \xrightarrow{v\gg 1} -\left(\frac{m}{v}\right)^2 \frac{1}{v^{s-1}} = -q^2v^{(s-1)-2},
\ee
so that
\begin{align}
\sigma^2_{U^{(L)}} &= v \sigma^2_{U^{(L-1)}} \left( \sigma^2_N - \sigma^2_{if} \right) + v\avg{U^{(L-1)}}^2\left(\sigma_N^2 + (v-1)\sigma^2_{if}\right) \nonumber \\
&\xrightarrow{v\gg 1} \sigma^2_{U^{(L-1)}} qv^{(s-1)} + \sigma^2_{U^{(L-1)}} q^2 v^{(s-1)-1} + q^{2L-1}(1-q) v^{(2L-1)(s-1)-2}
\end{align}
The second of the three terms is always subleading with respect to the first, so we can discard it for now. It remains to compare the first and the third terms. For $L\,{=}\,2$, since $\sigma^2_{U^{(1)}}\,{=}\,\sigma^2_N$, the first term depends on $v$ as $v^{2(s-1)-1}$, whereas the third is proportional to $v^{3(s-1)-2}$. For $L\,{\geq}\,3$ the dominant scaling is that of the third term only: for $L\,{=}\,3$ it can be shown by simply plugging the $L\,{=}\,2$ result into the recursion, and for larger $L$ it follows from the fact that replacing $\sigma^2_{U^{(L-1)}}$ in the first term with the third term of the precious step always yields a subdominant contribution. Therefore,
\be
\sigma^2_{U^{(L)}} \xrightarrow{v\gg 1}\left\lbrace
\begin{aligned}
& q^2 v^{2(s-1)-1} + q^3(1-q)v^{3(s-1)-2},&\text{ for }L=2,\\
& q^{2L-1}(1-q)v^{(2L-1)(s-1)-2},&\text{ for }L\geq 3.
\end{aligned}
\right.
\ee
Upon dividing the variance by the squared mean we get
\be
\frac{\sigma^2_{U^{(L)}}}{\avg{U^{(L)}}^2} \xrightarrow{v\gg 1}\left\lbrace
\begin{aligned}
& \frac{1}{q^2} \frac{1}{v^{2(s-1)-1}} + \frac{1-q}{q}\frac{1}{v^{(s-1)}},&\text{ for }L=2,\\
& \frac{1-q}{q}\frac{1}{v^{(s-1)}},&\text{ for }L\geq 3,
\end{aligned}
\right.
\ee
whose convergence to $0$ guarantees the concentration of the $U$'s around the average over all instances of the RHM.

\subsubsection{Statistics of the denominator $D$}

Here we compute the first and second moments of the denominator of $f^{(1\to L)}_{i_{1\to L}}(\mu_1;\alpha)$, 
\be\label{eq:den-def}
D^{(1\to L)}_{i_{1\to L}}(\mu_1) = \displaystyle\sum_{\mu_2, \dots, \mu_{L}=1}^v\sum_{\mu_{L+1}=1}^{n_c} N^{(1)}_{i_1}(\mu_1;\mu_2) \times\dots\times  N^{(L)}_{i_L}(\mu_{L};\mu_{L+1})
\ee

\paragraph{Level 1 $L\,{=}\,1$.} For $L\,{=}\,1$, $D$ is simply the sum over classes of the occurrences of a single production rule, $D^{(1)}\,{=}\,\sum_{\alpha} N_i(\mu_1;\alpha)$,
\begin{align}
\avg{D^{(1)}} &= n_c \frac{m}{v};\\
\sigma^2_{D^{(1)}}:=\var{D^{(1)}} &= n_c\sigma^2_N + n_c(n_c-1)c_{if} = n_c \left(\frac{m}{v}\right)^2 \frac{v-1}{v^s-1}\left(\frac{v^s}{m}-n_c\right)\nonumber\\&\xrightarrow{v\gg 1} n_c\left(\frac{m}{v}\right)^2\left(\frac{v}{m}-\frac{n_c}{v^{s-1}}\right);\\
c_{D^{(1)}}:=\cov{D^{(1)}(\mu_1), D^{(1)}(\nu_0)} &= -\frac{\var{D^{(1)}}}{(v-1)} = n_c c_N + n_c(n_c-1)c_{g},
\end{align}
where, in the last line, we used the identities $\sigma^2_N+(v-1)c_N\,{=}\,0$ from~\autoref{eq:cov-interclass} and $c_{if} + (v-1)c_{g}\,{=}\,0$ from~\autoref{eq:cov-general}.

\paragraph{Level 2 $L\,{=}\,2$.} For $L\,{=}\,2$,
\be
D^{(1\to 2)}_{i_{1\to 2}}(\mu_1) = \displaystyle\sum_{\mu_2}^v \sum_{\mu_3=1}^{n_c} N^{(1)}_{i_1}(\mu_1;\mu_2) \times N^{(2)}_{i_2}(\mu_2;\mu_3) = \sum_{\mu_2=1}^v N^{(1)}_{i_1}(\mu_1;\mu_2) D^{(2)}_{i_2}(\mu_2).
\ee
Therefore,
\begin{align}
\avg{D^{(1\to 2)}} &= v\left(\frac{m}{v}\right) \times \avg{D^{(1)}} =  \frac{n_c}{v}m
^2;\\
\sigma^2_{D^{(2)}}:=\var{D^{(1\to 2)}} &= \sum_{\mu_2,\nu_1=1}^v \left(\avg{N^{(1)}(\mu_1;\mu_2) N^{(1)}(\mu_1;\nu_1)}\avg{D^{(2)}(\mu_2)D^{(2)}(\nu_1)}-\avg{N}^2\avg{D^{(1)}}^2\right)\nonumber \\
&= \sum_{\mu_2, \nu_1=\mu_2} \dots + \sum_{\mu_2}\sum_{\nu_1 \neq \mu_2} \dots \nonumber \\
& = v\left(\sigma^2_N \sigma^2_{D^{(1)}} + \sigma^2_N \avg{D^{(1)}}^2 + \sigma^2_{D^{(1)}} \avg{N}^2  \right) + v(v-1)\left( c_{if} c_{D^{(1)}} + c_{if} \avg{D^{(1)}}^2 + c_{D^{(1)}} \avg{N}^2\right) \nonumber \\
& = v\left( \sigma^2_N \sigma^2_{D^{(1)}} +(v-1)c_{if} c_{D^{(1)}} \right) + v\avg{D^{(1)}}^2\left(\sigma_N^2 + (v-1)c_{if}\right) + v\avg{N}^2\left(\sigma_{D^{(1)}}^2 + (v-1)c_{D^{(1)}}\right) \nonumber \\
&= v \sigma^2_{D^{(1)}} \left( \sigma^2_N - c_{if} \right) + v\avg{D^{(1)}}^2\left(\sigma_N^2 + (v-1)c_{if}\right), \\
c_{D^{(2)}} &= -\frac{\sigma^2_{D^{(2)}}}{(v-1)}.
\end{align}

\paragraph{Level L.} In general,
\be
D^{(1\to L)}_{i_{1\to L}}(\mu_1) = \displaystyle\sum_{\mu_2=1}^v N^{(1)}_{i_1}(\mu_1;\mu_2) D^{(2\to L)}_{i_{2\to L}}(\mu_2).
\ee
Therefore,
\begin{align}\label{eq:den-stat-L}
\avg{D^{(L)}} &= v\left(\frac{m}{v}\right)\times \avg{D^{(L-1)}} = \frac{n_c}{v}m^L;\\
\sigma^2_{D^{(L)}} &= \sum_{\mu_2,\nu_1=1}^v \left(\avg{N^{(1)}(\mu_1;\mu_2) N^{(1)}(\mu_1;\nu_1)}\avg{D^{(2\to L)}(\mu_2;\alpha)D^{(2\to L)}(\nu_1;\alpha)}-\avg{N}^2\avg{D^{(1\to (L-1))}}^2\right)\nonumber \\
&= \sum_{\mu_2, \nu_1=\mu_2} \dots + \sum_{\mu_2}\sum_{\nu_1 \neq \mu_2} \dots \nonumber \\
& = v\left(\sigma^2_N \sigma^2_{D^{(L-1)}} + \sigma^2_N \avg{D^{(L-1)}}^2 + \sigma^2_{D^{(L-1)}} \avg{N}^2  \right) + v(v-1)\left( c_{if} c_{D^{(L-1)}} + c_{if} \avg{D^{(L-1)}}^2 + c_{D^{(L-1)}} \avg{N}^2\right) \nonumber \\
&= v \sigma^2_{D^{(L-1)}} \left( \sigma^2_N - c_{if} \right) + v\avg{D^{(L-1)}}^2\left(\sigma_N^2 + (v-1)c_{if}\right), \\
c_{D^{(L)}} &= -\frac{\sigma^2_{D^{(L)}}}{(v-1)}.
\end{align}

\paragraph{Concentration for large $m$.} Since the $D$'s can be expressed as a sum of different $U$'s, their concentration for $m\,{\gg}\,1$ follows directly from that of the $U$'s.

\subsubsection{Estimate of the conditional class probability}

We can now turn back to the original problem of estimating
\be
f^{(1\to L)}_{i_{1\to L}}(\alpha|\mu_1) = \frac{\displaystyle\sum_{\mu_2, \dots, \mu_{L}=1}^v N^{(1)}_{i_1}(\mu_1;\mu_2) \times\dots\times  N^{(L)}_{i_L}(\mu_{L};\alpha)}{\displaystyle\sum_{\mu_2, \dots, \mu_{L}=1}^v\sum_{\mu_{L+1}=1}^{n_c} N^{(1)}_{i_1}(\mu_1;\mu_2) \times\dots\times  N^{(L)}_{i_L}(\mu_{L};\mu_{L+1})} = \frac{U^{(1\to L)}_{i_{1\to L}}(\mu_1;\alpha)}{D^{(1\to L)}_{i_{1\to L}}(\mu_1)}.
\ee
Having shown that both numerator and denominator converge to their average for large $m$, we can expand for small fluctuations around these averages and write
\begin{align}\label{eq:cond-freq-exp}
f^{(1\to L)}_{i_{1\to L}}(\alpha|\mu_1) &= \frac{v^{-1}m^L\left(1+\frac{U^{(1\to L)}_{i_{1\to L}}(\mu_1;\alpha)-m^L/v}{m^L/v}\right)}{n_c v^{-1} m^L \left(1+\frac{D^{(1\to L)}_{i_{1\to L}}(\mu_1)-n_c m^L / v}{m^L}\right)} \\
&= \frac{1}{n_c} + \frac{1}{n_c}\frac{U^{(1\to L)}_{i_{1\to L}}(\mu_1;\alpha)-m^L/v}{m^L/v} - \frac{1}{n_c}\frac{D^{(1\to L)}_{i_{1\to L}}(\mu_1)-n_c m^L / v}{m^L / v} \nonumber \\
&= \frac{1}{n_c} + \frac{v}{n_c m^L}\left(U^{(1\to L)}_{i_{1\to L}}(\mu_1;\alpha)-\frac{1}{n_c}D^{(1\to L)}_{i_{1\to L}}(\mu_1)\right).
\end{align}
Since the conditional frequencies average to $n_c^{-1}$, the term in brackets averages to zero. We can then estimate the size of the fluctuations of the conditional frequencies (i.e. the `signal') with the standard deviation of the term in brackets.

It is important to notice that, for each $L$ and position $i_{1\to L}$, $D$ is the sum over $\alpha$ of $U$, and the $U$ with different $\alpha$ at fixed low-level feature $\mu_1$ are identically distributed. In general, for a sequence of identically distributed variables $(X_{\alpha})_{\alpha=1,\dots,n_c}$,
\be
\avg{\left(\frac{1}{n_c}\displaystyle\sum_{\beta=1}^v X_{\beta}\right)^2} = \frac{1}{n_c^2} \sum_{\beta=1}^{n_c} \left( \avg{X_\beta}^2 + \sum_{\beta'\neq \beta} \avg{X_\beta X_{\beta'}}\right) = \frac{1}{n_c}\left( \avg{X_\beta}^2 + \sum_{\beta'\neq \beta} \avg{X_\beta X_{\beta'}}\right).
\ee
Hence,
\begin{align}
\avg{\left(X_{\alpha}-\frac{1}{n_c}\displaystyle\sum_{\beta=1}^{n_c} X_{\beta}\right)^2} &= \avg{X_\alpha^2} + n_c^{-2} \sum_{\beta,\gamma=1}^{n_c} \avg{X_\beta X_\gamma} -2n_c^{-1} \sum_{\beta=1}^{n_c} \avg{X_\alpha X_\beta} \nonumber \\
&= \avg{X_\alpha^2} - n_c^{-1}\left( \avg{X_\alpha}^2 + \sum_{\beta\neq \alpha} \avg{X_\alpha X_{\beta}}\right) \nonumber \\
&= \avg{X_\alpha^2} - n_c^{-2}\avg{\left(\displaystyle\sum_{\beta=1}^{n_c} X_{\beta}\right)^2}.
\end{align}
In our case
\begin{align}
\avg{\left(U^{(1\to L)}_{i_{1\to L}}(\mu_1;\alpha)-\frac{1}{n_c}D^{(1\to L)}_{i_{1\to L}}(\mu_1)\right)^2} &= \avg{\left(U^{(1\to L)}_{i_{1\to L}}(\mu_1;\alpha)\right)^2} - {n_c}^{-2}\avg{\left(D^{(1\to L)}_{i_{1\to L}}(\mu_1)\right)^2}\nonumber\\
&= \sigma^2_{U^{(L)}} - n_c^{-2}
\sigma^2_{D^{(L)}},
\end{align}
where, in the second line, we have used that $\avg{U^{(L)}}\,{=}\, \avg{D^{(L)}}/n_c$ to convert the difference of second moments into a difference of variances. By~\autoref{eq:num-stat-L} and~\autoref{eq:den-stat-L},
\begin{align}
\sigma^2_{U^{(L)}} - n_c^{-2}
\sigma^2_{D^{(L)}} &= v \sigma^2_{U^{(L-1)}} \left( \sigma^2_N - \sigma^2_{if} \right) + v\avg{U^{(L-1)}}^2\left(\sigma_N^2 + (v-1)\sigma^2_{if}\right) \nonumber \\ &- \frac{v}{n_c^2} \sigma^2_{D^{(L-1)}} \left( \sigma^2_N - \sigma^2_{if} \right) - \frac{v}{n_c^2}\avg{D^{(L-1)}}^2\left(\sigma_N^2 + (v-1)\sigma^2_{if}\right) \nonumber \\
&= v\left( \sigma^2_N - \sigma^2_{if} \right) \left(\sigma^2_{U^{(L-1)}}-n_c^{-2}\sigma^2_{D^{(L-1)}}\right),
\end{align}
having used again that $\avg{U^{(L)}}\,{=}\, \avg{D^{(L)}}/n_c$. Iterating,
\be
\sigma^2_{U^{(L)}} - n_c^{-2}
\sigma^2_{D^{(L)}} = \left[v\left( \sigma^2_N - \sigma^2_{if} \right) \right]^{L-1} \left(\left(\sigma^2_{U^{(1)}}-n_c^{-2}\sigma^2_{D^{(1)}}\right)\right).
\ee
Since
\begin{align}
\sigma^2_{U^{(1)}} &= \frac{m}{v}\frac{v-1}{v}\frac{v^s-m}{v^s-1}\xrightarrow{v\gg 1} \frac{m}{v}, \nonumber\\
n_c^{-2}\sigma^2_{D^{(1)}} &= n_c^{-1}\sigma^2_N + n_c^{-1}(n_c-1)\sigma^2_{if}\xrightarrow{v\gg 1} n_c^{-1}\left(\frac{m}{v}\right)^2\left(\frac{v}{m}-\frac{n_c}{v^{s-1}}\right) = \frac{1}{n_c}\frac{m}{v}\left(1-\frac{m n_c}{v^s}\right),
\end{align}
One has
\be
\sigma^2_{U^{(L)}} - n_c^{-2}
\sigma^2_{D^{(L)}} \xrightarrow{v\gg 1} \frac{m^L}{v}\left(1-\frac{1-n_c m /v^s}{n_c}\right),
\ee
so that
\begin{align}\label{eq:signal}
\var{f^{(1\to L)}_{i_{1\to L}}(\alpha|\mu_1)} &= v^2 \frac{\avg{\left(U^{(1\to L)}_{i_{1\to L}}(\mu_1;\alpha)-\frac{1}{n_c}D^{(1\to L)}_{i_{1\to L}}(\mu_1)\right)^2}}{n_c^2 m^{2L}} \xrightarrow{v,n_c \gg 1} \frac{v}{n_c}\frac{1}{n_c m^L}.
\end{align}
\end{widetext}

\subsection{Introducing sampling noise due to the finite training set} In a supervised learning setting where only $P$ of the total data are available, the occurrences $N$ are replaced with their empirical counterparts $\hat{N}$. In particular, the empirical joint occurrence $\hat{N}(\mu;\alpha)$  (where we dropped level and positional indices to ease notation) coincides with the number of successes when sampling $P$ points without replacement from a population of $P_{\max}$ where only $N(\mu;\alpha)$ belong to class $\alpha$ and display feature $\mu$ in position $j$. Thus, $\hat{N}(\mu;\alpha)$ obeys a hypergeometric distribution where $P$ plays the role of the number of trials, $P_{\max}$ the population size, and the true occurrence $N(\mu;\alpha)$ the number of favorable cases. If $P$ is large and $P_{\max}$, $N(\mu;\alpha)$ are both larger than $P$, then
\be\label{eq:emp-joint}
\hat{N}(\mu;\alpha) \to \mathcal{N}\left(P\frac{N(\mu;\alpha)}{P_{\max}}, P\frac{N(\mu;\alpha)}{P_{\max}}\left(1-\frac{N(\mu;\alpha)}{P_{\max}}\right) \right),
\ee
where the convergence is meant as a convergence in probability and $\mathcal{N}(a,b)$ denotes a Gaussian distribution with mean $a$ and variance $b$. The statement above holds when the ratio $N(\mu;\alpha)/P_{\max}$ is away from $0$ and $1$, which is true with probability $1$ for large $v$ due to the concentration of $f(\alpha|\mu)$.
In complete analogy, the empirical occurrence $\hat{N}(\mu)$ obeys
\be\label{eq:emp-feat}
\hat{N}(\mu) \to \mathcal{N}\left(P\frac{N(\mu)}{P_{\max}}, P\frac{N(\mu)}{P_{\max}}\left(1-\frac{N(\mu)}{P_{\max}}\right) \right).
\ee
We obtain the empirical conditional frequency by the ratio of~\autoref{eq:emp-joint} and~\autoref{eq:emp-feat}. Since $N(\mu)\,{=}\,P_{\max}/v$ and $f(\alpha|\mu)\,{=}\,N(\mu;\alpha)/N(\mu)$, we have
\be
\hat{f}(\alpha|\mu) = \frac{\frac{f(\alpha|\mu)}{v} + \xi_P\sqrt{\frac{1}{P}\frac{f(\alpha|\mu)}{v}\left(1-\frac{f(\alpha|\mu)}{v}\right)}}{\frac{1}{v} + \zeta_P\sqrt{\frac{1}{P}\frac{1}{v}\left(1-\frac{1}{v}\right)}},
\ee
where $\xi_P$ and $\zeta_P$ are correlated zero-mean and unit-variance Gaussian random variables over independent drawings of the $P$ training points. By expanding the denominator of the right-hand side for large $P$ we get, after some algebra,
\begin{widetext}
\be\label{eq:emp-freq}
\hat{f}(\alpha|\mu) \simeq f(\alpha|\mu) +\xi_P \sqrt{\frac{v f(\alpha|\mu)}{P} \left(1-\frac{f(\alpha|\mu)}{v}\right)} - \zeta_P f(\alpha|\mu) \sqrt{\frac{v}{P}\left(1-\frac{1}{v}\right)}.
\ee
\end{widetext}
Recall that, in the limit of large $n_c$ and $m$, $f(\alpha|\mu)=n_c^{-1}(1 + \sigma_f \xi_{\text{RHM}})$ where $\xi_{\text{RHM}}$ is a zero-mean and unit-variance Gaussian variable over the realizations of the RHM, while $\sigma_f$ is the `signal', $\sigma^2_f\,{=}\,v / m^L$ by~\autoref{eq:signal}. As a result,
\be\label{eq:emp-freq-limit}
\hat{f}(\alpha|\mu) \xrightarrow{n_c,m, P\gg 1} \frac{1}{n_c}\left(1 + \sqrt{\frac{v}{m^L}}\xi_{\text{RHM}} + \sqrt{\frac{v n_c}{P}}\xi_P \right).
\ee
\subsection{Sample complexity} From~\autoref{eq:emp-freq-limit} it is clear that for the `signal' $\hat{f}$, the fluctuations due to noise must be smaller than those due to the random choice of the composition rules. Therefore, the crossover takes place when the two nose terms have the same size, occurring at $P\,{=}\,P_c$ such that
\be
\sqrt{\frac{v}{m^L}} =  \sqrt{\frac{v n_c}{P_c}} \Rightarrow P_c = n_c m^L.
\ee

\section{Improved Sample Complexity via Clustering}\label{app:clustering}

In this section, we consider the maximal dataset case $n_c\,{=}\,v$ and $m\,{=}\,v^{s-1}$, and show that a distance-based clustering method acting on the hidden representations of~\autoref{eq:rep-gradient-proof} would identify synonyms at $P\,{\simeq}\,\sqrt{n_c}m^L$. Let us then imagine feeding the representations updates $\Delta f_h(\bm{\mu})$ of~\autoref{eq:rep-gradient-proof} to a clustering algorithm aimed at identifying synonyms. This algorithm is based on the distance between the representations of different tuples of input features $\bm{\mu}$ and $\bm{\nu}$,
\begin{align}
\lVert \Delta f(\bm{\mu}) - \Delta f(\bm{\nu})  \rVert^2 := \frac{1}{H}\sum_{h=1}^H \left(\Delta f_h(\bm{\mu})-\Delta f_h(\bm{\nu})\right)^2,
\end{align}
where $H$ is the number of hidden neurons. By defining
\begin{align}
\hat{g}_{\alpha}(\bm{\mu}):= \frac{\hat{N}_1(\bm{\mu};\alpha)}{P}-\frac{1}{n_c}\frac{\hat{N}_1(\bm{\mu})}{P},
\end{align}
and denoting with $\hat{\bm{g}}(\bm{\mu})$ the $n_c$-dimensional sequence having the $\hat{g}_{\alpha}$'s as components, we have
\begin{widetext}
\begin{align}
\lVert \Delta f(\bm{\mu}) - \Delta f(\bm{\nu})  \rVert^2 &= \sum_{\alpha,\beta=1}^{n_c} \left(\frac{1}{H}\sum_{h}^{H} a_{h,\alpha} a_{h,\beta}\right) \left(\hat{g}_\alpha(\bm{\mu})-\hat{g}_\alpha(\bm{\nu})\right)\left(\hat{g}_\beta(\bm{\mu})-\hat{g}_\beta(\bm{\nu})\right)\nonumber\\
&\xrightarrow{H\to\infty} \sum_{\alpha=1}^{n_c}\left(\hat{g}_\alpha(\bm{\mu})-\hat{g}_\alpha(\bm{\nu})\right)^2 = \lVert \hat{\bm{g}}(\bm{\mu})-\hat{\bm{g}}(\bm{\nu}) \rVert^2,
\end{align}
\end{widetext}
where we used the i.i.d. Gaussian initialization of the readout weights to replace the sum over neurons with $\delta_{\alpha,\beta}$. 

Due to the sampling noise, from~\autoref{eq:emp-joint} and~\autoref{eq:emp-feat}, when $1\,{\ll}\,P\,{\ll}\,P_{\text{max}}$,
\begin{align}
\hat{g}_{\alpha}(\bm{\mu}) = g_{\alpha}(\bm{\mu}) + \sqrt{\frac{1}{n_c m v P}}\,\eta_{\alpha}(\bm{\mu}),
\end{align}
where $\eta_{\alpha}(\bm{\mu})$ is a zero-mean and unit-variance Gaussian noise and $g$ without hat denotes the $P\to P_{\text{max}}$ limit of $\hat{g}$. In the limit $1\,{\ll}\,P\,{\ll}\,P_{\text{max}}$, the noises with different $\alpha$ and $\bm{\mu}$ are independent of each other. Thus,
\begin{align}\label{eq:rep-dist}
&\lVert \hat{\bm{g}}(\bm{\mu})-\hat{\bm{g}}(\bm{\nu}) \rVert^2 = \nonumber\\
&\lVert \bm{g}(\bm{\mu})-\bm{g}(\bm{\nu}) \rVert^2 + \frac{1}{n_c m v P}\lVert \bm{\eta}(\bm{\mu})-\bm{\eta}(\bm{\nu}) \rVert^2 +\nonumber\\
&\frac{2}{\sqrt{n_c m v P}} \left(\bm{g}(\bm{\mu})-\bm{g}(\bm{\nu})\right)\cdot\left(\bm{\eta}(\bm{\mu})-\bm{\eta}(\bm{\nu})\right).
\end{align}
If $\bm{\mu}$ and $\bm{\nu}$ are synonyms, then $\bm{g}(\bm{\mu})\,{=}\,\bm{g}(\bm{\nu})$ and only the noise term contributes to the right-hand side of~\autoref{eq:rep-dist}. If this noise is sufficiently small, then the distance above can be used to cluster tuples into synonymic groups. 

By the independence of the noises and the Central Limit Theorem, for $n_c\,{\gg}\,1$,
\begin{align}
\lVert \bm{\eta}(\bm{\mu})-\bm{\eta}(\bm{\nu}) \rVert^2 \sim \mathcal{N}(2 n_c, \mathcal{O}(\sqrt{n_c})),
\end{align}
over independent samplings of the $P$ training points. The $g$'s are also random variables over independent realizations of the RHM with zero mean and variance proportional to the variance of the conditional probabilities $f(\alpha|\bm{\mu})$ (see~\autoref{eq:cond-freq-exp} and~\autoref{eq:signal}),
\begin{align}
\var{g_\alpha(\bm{\mu})} = \frac{1}{n_c m v n_c m^L} = \frac{1}{n_c m v P_c}.
\end{align}
To estimate the size of $\lVert \bm{g}(\bm{\mu})-\bm{g}(\bm{\nu}) \rVert^2$ we must take into account the correlations (over RHM realizations) between $g$'s with different class label and tuples. However, in the maximal dataset case $n_c\,{=}\,v$ and $m\,{=}\,v^{s-1}$, both the sum over classes and the sum over tuples of input features of the joint occurrences $N(\bm{\mu};\alpha)$ are fixed deterministically. The constraints on the sums allow us to control the correlations between occurrences of the same tuple within different classes and of different tuples within the same class, so that the size of the term $\lVert \bm{g}(\bm{\mu})-\bm{g}(\bm{\nu}) \rVert^2$ for $n_c\,{=}\,v\,{\gg}\,1$ can be estimated via the Central Limit Theorem:
\begin{align}
\lVert \bm{g}(\bm{\mu})-\bm{g}(\bm{\nu}) \rVert^2 \sim \mathcal{N}\left( \frac{2 n_c}{n_c m v P_c}, \frac{\mathcal{O}(\sqrt{n_c})}{n_c m v P_c} \right).
\end{align}
The mixed term $\left(\bm{g}(\bm{\mu})-\bm{g}(\bm{\nu})\right)\cdot\left(\bm{\eta}(\bm{\mu})-\bm{\eta}(\bm{\nu})\right)$ has zero average (both with respect to training set sampling and RHM realizations) and can also be shown to lead to relative fluctuations of order $\mathcal{O}(\sqrt{n_c})$ in the maximal dataset case.

\begin{figure*}[h]
    \centering
    \includegraphics[width=\linewidth]{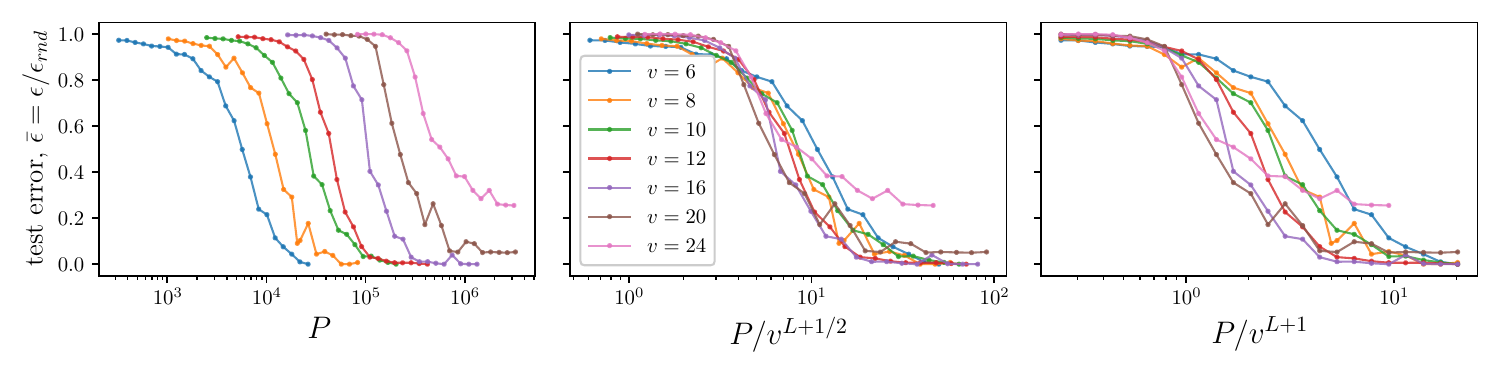}
    \caption{\textbf{Sample complexity for layerwise training, $m=n_c=v$, $L=3, s=2$.} Training of a $L$-layers network is performed layerwise by alternating one-step GD as described in Section 4.C and clustering of the hidden representations. Clustering of the $mv=v^2$ representations for the different one-hot-encoded input patches is performed with the $k$-means algorithms. Clustered representations are then orthogonalized and the result is given to the next one-step GD procedure. Left: Test error vs number of training points. Different colors correspond to different values of $v$. Center: collapse of the test error curves when rescaling the $x$-axis by $v^{L + 1/2}$. Right: analogous, when rescaling the $x$-axis by $v^{L + 1}$. The curves show a better collapse when rescaling by $v^{L + 1/2}$, suggesting that these layerwise algorithms have an advantage of a factor $\sqrt{v}$ over end-to-end training with deep CNNs, for which $P^* = v^{L+1}$.}
    \label{fig:terr_vs_p_layerwise}
\end{figure*}

Tu sum up, we have that, for synonyms,
\begin{align}
\lVert \hat{\bm{g}}(\bm{\mu})-\hat{\bm{g}}(\bm{\nu}) \rVert^2 &= \lVert \bm{\eta}(\bm{\mu})-\bm{\eta}(\bm{\nu}) \rVert^2\nonumber\\ &\sim \frac{1}{m v P}\left(1 + \frac{1}{\sqrt{n_c}}\xi_P\right),
\end{align}
where $\xi_P$ is some $\mathcal{O}(1)$ noise dependent on the training set sampling. If $\bm{\mu}$ and $\bm{\nu}$ are not synonyms, instead,
\begin{align}
\lVert \hat{\bm{g}}(\bm{\mu})-\hat{\bm{g}}(\bm{\nu}) \rVert^2 &\sim \frac{1}{m v P}\left(1 + \frac{1}{\sqrt{n_c}}\xi_P\right) \nonumber\\
&+ \frac{1}{m v P_c}\left(1 + \frac{1}{\sqrt{n_c}}\xi_{\text{RHM}}\right),
\end{align}
where $\xi_{\text{RHM}}$ is some $\mathcal{O}(1)$ noise dependent on the RHM realization. In this setting, the signal is the deterministic part of the difference between representations of non-synonymic tuples. Due to the sum over class labels, the signal is scaled up by a factor $n_c$, whereas the fluctuations (stemming from both sampling and model) are only increased by $\mathcal{O}\left(\sqrt{n_c}\right)$. Therefore, the signal required for clustering emerges from the sampling noise at $P\,{=}\, P_c/\sqrt{n_c}\,{=}\, \sqrt{n_c} m^L$, equal to $v^{1/2+L(s-1)}$ in the maximal dataset case. This prediction is tested for $s\,{=}\,2$ in~\autoref{fig:terr_vs_p_layerwise}, which shows the error achieved by a layerwise algorithm which alternates single GD steps to clustering of the resulting representations~\cite{malach2018provably, malach2020implications}. More specifically, the weights of the first hidden layer are updated with a single GD step while keeping all the other weights frozen. The resulting representations are then clustered, so as to identify groups of synonymic level-$1$ tuples. The centroids of the ensuing clusters, which correspond to level-$2$ features, are orthogonalized and used as inputs of another one-step GD protocol, which aims at identifying synonymic tuples of level-$2$ features. The procedure is iterated $L$ times.

\section{Intrinsic Dimensionality of Data Representations}
\label{app:intrinsic-dimension}
In deep learning, the representation of data at each layer of a network can be thought of as lying on a manifold in the layer's activation space. Measures of the \textit{intrinsic dimensionality} of these manifolds can provide insights into how the networks lower the dimensionality of the problem layer by layer. However, such measurements have challenges. One key challenge is that it assumes that real data exist on a smooth manifold, while in practice, the dimensionality is estimated based on a discrete set of points. This leads to counter-intuitive results such as an increase in the intrinsic dimensionality with depth, especially near the input. An effect that is impossible for continuous smooth manifolds. We resort to an example to illustrate how this increase with depth can result from spurious effects. Consider a manifold of a given intrinsic dimension  that undergoes a transformation where one of the coordinates is multiplied by a large factor. This operation would result in an elongated manifold that appears one-dimensional. The measured intrinsic dimensionality would consequently be one, despite the higher dimensionality of the manifold.
In the context of neural networks, a network that operates on such an elongated manifold could effectively 'reduce' this extra, spurious dimension. This could result in an increase in the observed intrinsic dimensionality as a function of network depth, even though the actual dimensionality of the manifold did not change.

\begin{figure*}[ht]
    \centering
    \includegraphics[width=.8\textwidth]{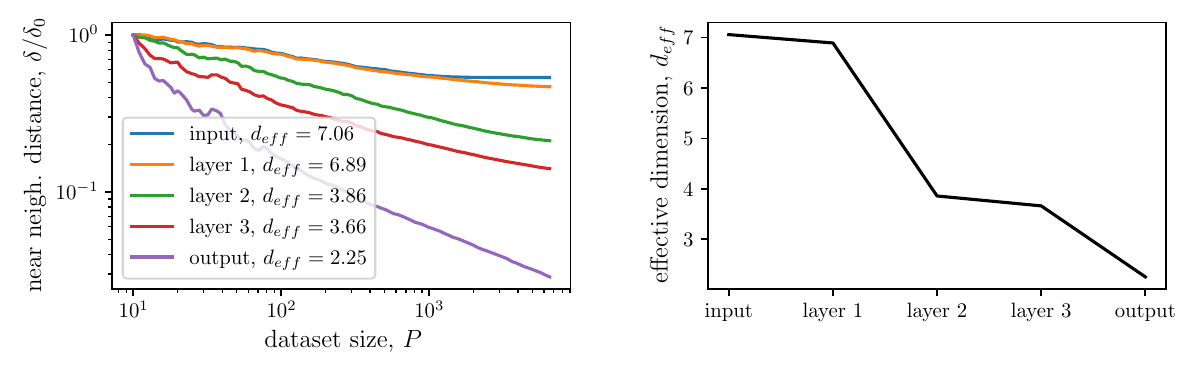}
    \caption{Effective dimension of the internal representation of a CNN trained on one instance of the RHM with $m=n_c=v, L=3$ resulting in $P_{\max} = 6'232$. Left: average nearest neighbor distance of input or network activations when probing them with a dataset of size $P$. The value reported on the $y$-axis is normalized by $\delta_0 = \delta(P=10)$. The slope of $\delta(P)$ is used as an estimate of the effective dimension. Right: effective dimension as a function of depth. We observe a monotonic decrease, consistent with the idea that the dimensionality of the problem is reduced by DNNs with depth.}
    \label{fig:deff}
\end{figure*}

In the specific case of our data, the intrinsic dimensionality of the internal representations of deep CNNs monotonically decreases with depth, see~\autoref{fig:deff}, consistently with the idea proposed in the main text that the CNNs solve the problem by reducing the effective dimensionality of data layer by layer. We attribute this monotonicity to the absence of spurious or noisy directions that might lead to the counter-intuitive effect described above.

\section{Additional Results on Sample Complexity}\label{app:add-figures}

\begin{figure*}[ht]
    \centering
    \includegraphics[width=\linewidth]{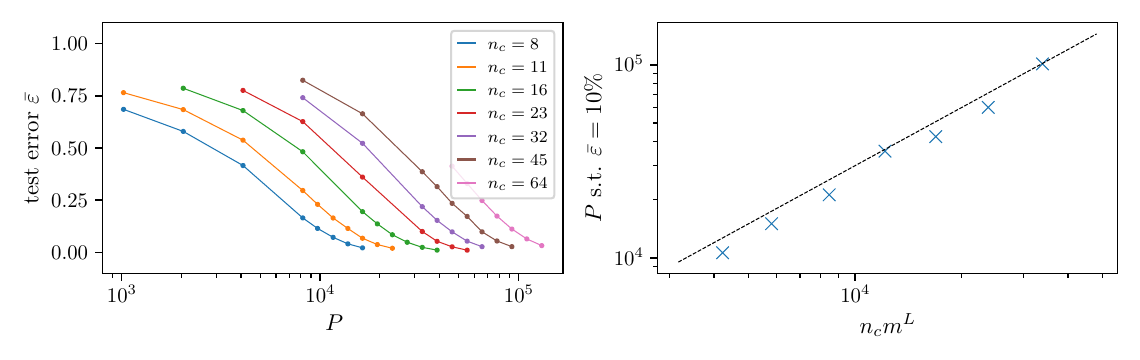}
    \caption{\textbf{Sample complexity of deep CNNs, for $L\,{=}\,s\,{=}\,2$, $v\,{=}\,256$, $m\,{=}\,23$ and different values of $n_c$.} Left: Test error vs number of training points with the color indicating the number of classes (see key). 
    Right: sample complexity $P^*$ (crosses) and law $P^*\,{=}\,n_c m^L$ (black dashed).}
    \label{fig:terr_vs_p_diff_nc}
\end{figure*}

\begin{figure*}[ht]
    \centering
    \includegraphics[width=\linewidth]{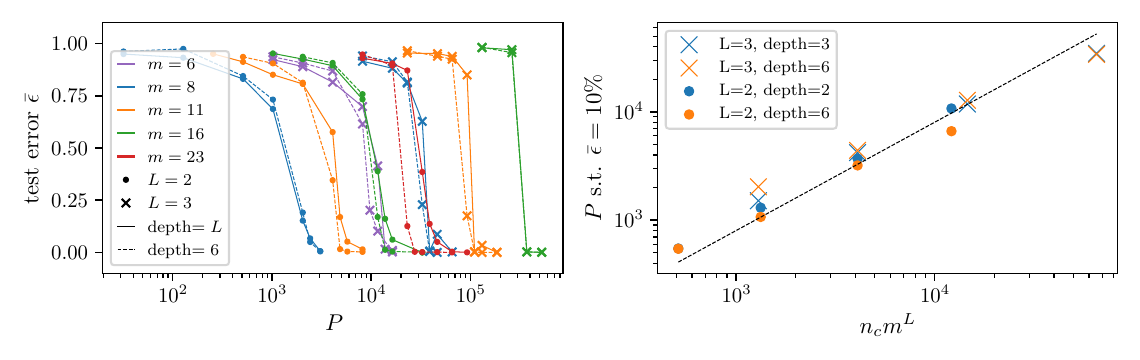}
    \caption{\textbf{Sample complexity of deep fully-connected networks with different depth, for $s\,{=}\,2$ and $m\,{=}\,n_c\,{=}\,v$.} Left: Test error vs number of training points. The color denotes the value of $m\,{=}\,n_c\,{=}\,v$, the marker the hierarchy depth of the RHM $L$. Solid lines represent networks having depth $L$, while dashed lines correspond to networks with depth $6\,{>}\,L$. Notice that, in all cases, the behavior of the test error is roughly independent of the network depth. 
    Right: sample complexity $P^*$ (crosses and circles). With respect to the case of deep CNNs tailored to the structure of the RHM, the sample complexity of generic deep networks seems to display an additional factor of $s^L$ independently of $n_c$, $m$, and $v$.}
    \label{fig:terr_vs_p_diff_arch}
\end{figure*}

\begin{figure*}[ht]
    \centering
    \includegraphics[width=.5\textwidth]{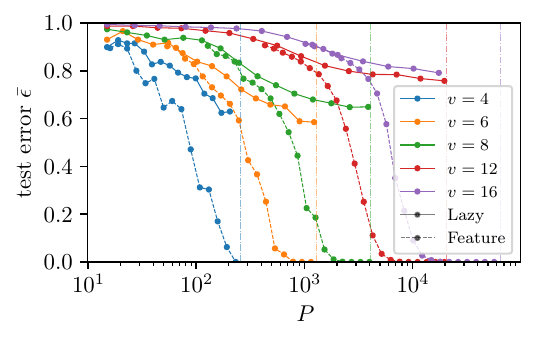}
    \caption{\textbf{Learning curves of depth-$(L+1)$ CNNs, for $L=2$, $s\,{=}\,2$ and $m\,{=}\,n_c\,{=}\,v$ trained in the `lazy' regime (full lines)---where they are equivalent to a kernel method \cite{jacot2018neural}---and in the `feature' learning regime (dashed lines).} Different colors correspond to different vocabulary sizes $v$. Vertical lines signal $P_{\rm max}=v^{s^L}$.}
    \label{fig:ntk}
\end{figure*}

\begin{figure*}[ht]
    \centering
    \includegraphics[width=.45\textwidth]{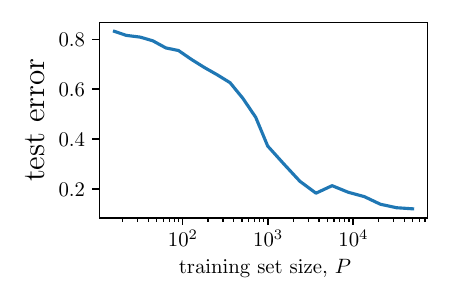}
    \caption{\textbf{Test error vs number of training points for a ResNet18 trained on subsamples of the CIFAR10 dataset.} Results are the average of 10 jointly different initializations of the networks and dataset sampling.}
    \label{fig:terr_vs_P_cifar10}
\end{figure*}

This section collects additional results on the sample complexity of deep networks trained on the RHM (\autoref{fig:terr_vs_p_diff_nc} and \autoref{fig:terr_vs_p_diff_arch}), on the learning curves for `lazy' neural networks (\autoref{fig:ntk}), and for a ResNet18 trained on different sub-samples of the benchmark dataset CIFAR10 (\autoref{fig:terr_vs_P_cifar10}).

\autoref{fig:terr_vs_p_diff_nc} shows the behavior of the sample complexity with varying number of classes $n_c$ when all the other parameters of the RHM are fixed, confirming the linear scaling discussed in the main text.

\autoref{fig:terr_vs_p_diff_arch} shows the behavior of the sample complexity for deep fully-connected networks having depth larger than $L+1$, which are not tailored to the structure of the RHM. Notice that changing architecture seems to induce an additional factor of $2^L$ to the sample complexity, independent of $v$, $n_c$ and $m$. This factor is also polynomial in the input dimension.  

\autoref{fig:ntk} presents the learning curves for deep CNNs tailored to the structure of the model and trained in the lazy regime on the maximal case, i.e., $n_c=v$ and $m=v^s$. In particular, we consider the infinite-width limit of CNNs with all layers scaled by a factor $H^{-1/2}$, including the last. In this limit, CNNs become equivalent to a kernel method~\cite{jacot2018neural}, with an architecture-dependent kernel known as the \textit{Neural Tangent Kernel} (NTK). In our experiments, we use the analytical form of this kernel (see, e.g.,~\cite{cagnetta2023can}) and train a kernel logistic regression classifier up to convergence. Our main result is that, in the lazy regime, the generalization error stays finite even when $P\approx P_{\rm max}$; thus, kernels suffer from the curse of dimensionality.

Notice that the learning curves of the lazy regime follow those of the feature learning regime for $P\ll P^*$. This is because the CNN kernel can also exploit local correlations between the label and input patches~\cite{cagnetta2023can} to improve its performance. However, unlike in the feature regime, kernels cannot build a hierarchical representation, and thus their test error does not converge to zero.

\end{document}